\documentclass[10pt,logo,copyright]{nvidiatechreport}
\usepackage[authoryear,sort&compress,round]{natbib}

\usepackage[utf8]{inputenc} 
\usepackage[T1]{fontenc}    
\usepackage{xcolor}
\usepackage{parskip}        
\usepackage{url}            
\usepackage{booktabs}       
\usepackage{amsfonts}       
\usepackage{nicefrac}       
\usepackage{microtype}      
\usepackage{xcolor}         
\usepackage[dvipsnames]{xcolor} 
\usepackage{graphicx}
\usepackage{animate}        
\usepackage{subcaption}
\usepackage{tabularx}
\usepackage{makecell}
\usepackage{adjustbox}
\usepackage{setspace}
\newcolumntype{M}[1]{>{\centering\arraybackslash}m{#1}}
\usepackage{float}
\usepackage{tikz}
\usetikzlibrary{positioning,shapes,arrows}
\usepackage{amsmath,amsfonts,bm, bbm,leftindex}
\usepackage{multirow}
\usepackage{comment}
\usepackage{gensymb}
\usepackage{lipsum}
\usetikzlibrary{arrows.meta, positioning, fit}
\usepackage[para]{threeparttable}
\usepackage{tikz}
\usetikzlibrary{tikzmark}
\usepackage{caption}
\def\eqref#1{equation~\ref{#1}}

\def\1{\bm{1}}

\def\x{\mathbf{x}}
\def\y{\mathbf{y}}

\DeclareMathAlphabet{\mathsfit}{\encodingdefault}{\sfdefault}{m}{sl}
\SetMathAlphabet{\mathsfit}{bold}{\encodingdefault}{\sfdefault}{bx}{n}

\makeatletter
\let\save@mathaccent\mathaccent
\newcommand*\if@single[3]{%
  \setbox0\hbox{${\mathaccent"0362{#1}}^H$}%
  \setbox2\hbox{${\mathaccent"0362{\kern0pt#1}}^H$}%
  \ifdim\ht0=\ht2 #3\else #2\fi
  }
\newcommand*\rel@kern[1]{\kern#1\dimexpr\macc@kerna}
\newcommand*\widebar[1]{\@ifnextchar^{{\wide@bar{#1}{0}}}{\wide@bar{#1}{1}}}
\newcommand*\wide@bar[2]{\if@single{#1}{\wide@bar@{#1}{#2}{1}}{\wide@bar@{#1}{#2}{2}}}
\newcommand*\wide@bar@[3]{%
  \begingroup
  \def\mathaccent##1##2{%
    \let\mathaccent\save@mathaccent
    \if#32 \let\macc@nucleus\first@char \fi
    \setbox\z@\hbox{$\macc@style{\macc@nucleus}_{}$}%
    \setbox\tw@\hbox{$\macc@style{\macc@nucleus}{}_{}$}%
    \dimen@\wd\tw@
    \advance\dimen@-\wd\z@
    \divide\dimen@ 3
    \@tempdima\wd\tw@
    \advance\@tempdima-\scriptspace
    \divide\@tempdima 10
    \advance\dimen@-\@tempdima
    \ifdim\dimen@>\z@ \dimen@0pt\fi
    \rel@kern{0.6}\kern-\dimen@
    \if#31
      \overline{\rel@kern{-0.6}\kern\dimen@\macc@nucleus\rel@kern{0.4}\kern\dimen@}%
      \advance\dimen@0.4\dimexpr\macc@kerna
      \let\final@kern#2%
      \ifdim\dimen@<\z@ \let\final@kern1\fi
      \if\final@kern1 \kern-\dimen@\fi
    \else
      \overline{\rel@kern{-0.6}\kern\dimen@#1}%
    \fi
  }%
  \macc@depth\@ne
  \let\math@bgroup\@empty \let\math@egroup\macc@set@skewchar
  \mathsurround\z@ \frozen@everymath{\mathgroup\macc@group\relax}%
  \macc@set@skewchar\relax
  \let\mathaccentV\macc@nested@a
  \if#31
    \macc@nested@a\relax111{#1}%
  \else
    \def\gobble@till@marker##1\endmarker{}%
    \futurelet\first@char\gobble@till@marker#1\endmarker
    \ifcat\noexpand\first@char A\else
      \def\first@char{}%
    \fi
    \macc@nested@a\relax111{\first@char}%
  \fi
  \endgroup
}
\makeatother

\hypersetup{pdftitle={Industrial ReMEmbR: Scaling Spatiotemporal Memory to Multi-Agent Robot Fleets}, pdfauthor={Idealworks and NVIDIA}}

\usepackage{booktabs}
\usepackage{tikz}
\usetikzlibrary{arrows.meta,positioning,fit,backgrounds,calc,shapes.geometric}
\usepackage{pgfplots}
\pgfplotsset{compat=1.18}
\providecolor{nvidiagreen}{HTML}{76B900}

\usepackage{xspace}

\newsavebox{\promptsavebox}
\newenvironment{promptbox}[1]{%
    \par\medskip\noindent\footnotesize\fboxsep=5pt\raggedright\textbf{#1}\par\nopagebreak\vspace{2pt}%
    \begin{lrbox}{\promptsavebox}\begin{minipage}{\dimexpr\linewidth-2\fboxsep-2\fboxrule\relax}%
    \ttfamily\raggedright\sloppy\setlength{\parskip}{3pt}%
}{%
    \end{minipage}\end{lrbox}\noindent\fbox{\usebox{\promptsavebox}}\par\medskip%
}
\newcommand{\stackRemembr}{\mbox{Upgraded ReMEmbR}\xspace}
\newcommand{\industrialRemembr}{\mbox{Industrial ReMEmbR}\xspace}

\newcommand{\FIGREF}[1]{Fig.~\ref{#1}} 
\newcommand{\TABREF}[1]{Tab.~\ref{#1}}
\definecolor{darkred}{rgb}{0.7, 0.0, 0.0}

\usepackage{pifont}

\usepackage[nameinlink]{cleveref}
\crefname{equation}{Eq.}{Eqs.}
\crefname{figure}{Fig.}{Figs.}
\crefname{section}{Sec.}{Sec.}
\crefname{appendix}{App.}{App.}
\crefname{table}{Tab.}{Tabs.}
\crefname{algorithm}{Algo}{Algo}
\crefname{thm}{Thm}{Thm}
\Crefname{thm}{Thm}{Thm}
\crefname{prop}{Prop}{Prop}

\newcommand{\crefnames}[3]{%
  \@for\next:=#1\do{%
    \expandafter\crefname\expandafter{\next}{#2}{#3}%
  }%
}

\title{\industrialRemembr: Scaling Spatiotemporal Memory to Robot Fleets}

\author[1]{Jad Yammine}
\author[1]{Malek Wahidi}
\author[2]{Abrar Anwar}
\author[2]{Yan Chang}
\author[2]{Mihir Acharya}
\author[1]{Anthony Rizk}

\affil[1]{Idealworks}
\affil[2]{NVIDIA}

\begin{abstract}
Warehouse robot fleets continuously observe operational conditions, but preserving these observations as reusable semantic memory incurs substantial computation, communication, and storage costs. For example, explaining recurring aisle blockages can require reviewing video from multiple robots across several shifts. We present \industrialRemembr{}, a shared spatiotemporal memory system that extends NVIDIA ReMEmbR to robot fleets and reduces server-side computation by generating scene descriptions only on demand. On each robot, an NVIDIA Jetson encodes video clips into retrieval embeddings with Cosmos-Embed1, vision tokens with Cosmos3-Edge, and reconstruction latents with the Wan2.2 VAE, without uploading raw camera frames. The encoding models are re-used to detect people in image patches, zeroing them out to support PII abstraction for GDPR-compliant deployments.  We split Cosmos3-Edge at its vision–language boundary, placing the vision encoder on the robot and the language decoder on an on-premises server. A Nemotron agent retrieves memories by content, position, and time and queries the decoder using cached vision tokens. This design eliminates caption generation at write time, while cached-token decoding reproduces the unsplit model’s output in all 450 tested decodes. On the NaVQA benchmark, the updated model stack achieves comparable retrieval recall while reducing unparsable answers from 4.5\% to 1.9\% with an agent half the parameter size. We further demonstrate a live semantic traffic map that classifies aisle conditions directly from retrieval embeddings. Across 914 simulated warehouse frames, this method exceeds the best caption-based alternative by 6.0 percentage points. By decoupling memory encoding from description, Industrial ReMEmbR transforms a fleet’s collective experience into an extensible foundation for spatial and temporal reasoning, including faithful scene reconstruction, context-guided elaboration, and semantic mapping of dynamic warehouse conditions as they unfold.
\end{abstract}

\begin{document}
\maketitle
\abscontent

\section{Introduction}
\label{sec:intro}

An embodied memory lets a robot recall what it has seen, where, and when, long after the moment has passed. NVIDIA's ReMEmbR~\citep{anwar2025remembr}, a spatiotemporal memory system designed for long-horizon visual question answering for robot navigation, showed that a single robot can build such a memory from its own camera stream. On a warehouse floor, however, that robot is only one of many. A fleet management system, or fleet manager, continuously tracks dozens of autonomous mobile robots (AMRs) and other integrated devices in a modern warehouse, assigning them orders and planning their routes over VDA5050, the standard interface between mobile robots and fleet management systems. Obstacles outside the tracking system are left to each robot's intelligent navigation system to recognize and avoid. What a passing robot sees is rarely kept, because continuously uploading and storing raw video from every robot is expensive, so the rich semantic context of each scene never reaches the rest of the fleet.

To illustrate, consider the case of a fleet management system that keeps routing AMRs through an aisle partially blocked by a clutter of untracked obstacles. Each robot independently maneuvers around the clutter, losing time before carrying on with its scheduled tasks. A few days later, a human operator reviewing the daily logs notices that task completion times slipped that day, but has no record of the scene from which to trace its root cause. Neither the fleet manager nor the operator has the semantic context that could have prevented the congestion, by adaptively routing later robots away from the aisle, or its recurrence, by revealing what caused it.

Closing this contextual blind spot requires scaling embodied memory from a single robot to the entire fleet, with every connected robot acting as a live reporter from the warehouse floor that continuously contributes its observations, tagged with location and time, to a shared fleet memory. In the scenario above, the first robot to encounter the clutter would report it on a live semantic map, letting the fleet manager route subsequent robots around the blocked aisle. Days later, the operator could ask the fleet's shared semantic memory what caused the reroutes and receive an answer grounded in direct visual observation: when and where the clutter appeared, and what it consisted of. The same record can supply the world state of a centralized planner such as RobotFleet~\citep{gupta2025robotfleet}, which schedules heterogeneous robots from declarative statements about the site and replans when a robot reports a discovery.

To this end, we build on ReMEmbR, which captions each short video clip, stores the caption along with the robot's time and location and a corresponding caption embedding, and lets an agentic model perform semantic retrieval over the vector database to answer arbitrary questions. Scaling this design from one robot to a fleet breaks three of its assumptions, and each break motivates one principle of our design. First, ReMEmbR generates a caption for every clip as it is written, so its cost grows with every robot added and every hour it operates, and continuously uploading raw video from every robot is costly to begin with. We therefore \emph{encode once}, compressing each clip on the robot's own edge compute into compact representations that every later step reuses. Second, ReMEmbR answers occasional questions, whereas the fleet manager also needs a continuously updated view of the site. We therefore \emph{interpret continuously}, reading that state directly from each memory as it arrives, without generating any text. Third, the questions a memory must answer are not known when it is first encoded. We therefore \emph{elaborate on demand}, deferring any description of a scene until a question about it arrives. In the blocked aisle example, the clutter is encoded once by the first robot that passes it, flagged on the live map as soon as its memory arrives, and described in detail only when the operator asks about it days later.

\begin{figure}[t]
    \centering
    \resizebox{\linewidth}{!}{%
        \begin{tikzpicture}[
            font=\sffamily\footnotesize,
            lane/.style={draw=black!30, rounded corners=6pt, fill=black!1.5, line width=0.6pt},
            lanetitle/.style={anchor=north west, font=\sffamily\bfseries\footnotesize},
            component/.style={draw=black!25, fill=black!4, rounded corners=3pt, text width=2.9cm, minimum height=1.0cm, align=center, inner sep=3pt},
            model/.style={component, draw=nvidiagreen!75!black, fill=nvidiagreen!10},
            memory/.style={component, draw=blue!35!black, fill=blue!5},
            actor/.style={component, fill=white, draw=black!45},
            source/.style={component, fill=white},
            flow/.style={-{Stealth[length=2.1mm]}, draw=black!65, line width=0.7pt},
            bidir/.style={{Stealth[length=2.1mm]}-{Stealth[length=2.1mm]}, draw=black!65, line width=0.7pt},
            ref/.style={draw=blue!35!black, dashed, line width=0.6pt},
            note/.style={font=\sffamily\scriptsize, text=black!65, fill=white, inner sep=1.5pt},
            badge/.style={circle, fill=nvidiagreen!70!black, text=white, font=\sffamily\bfseries\scriptsize, inner sep=0pt, minimum size=4.4mm},
            traffic/.style={fill=orange!70!yellow!55},
            clear/.style={fill=blue!45!cyan!40},
            robot/.pic={\fill[black!78] (0.105,0) -- (-0.07,0.07) -- (-0.037,0) -- (-0.07,-0.07) -- cycle;}
        ]
            \draw[lane, fill=black!0.5] (0.70,9.90) rectangle (19.45,12.80);
            \draw[lane, fill=black!1] (0.55,10.05) rectangle (19.30,12.95);
            \draw[lane] (0.40,10.20) rectangle (19.15,13.10);
            \node[badge] at (0.80,12.72) {1};
            \node[lanetitle] at (1.08,12.98) {Encode once\enspace\textmd{\color{black!55}on every robot (edge compute), $\times N$}};

            \node[source] (clip) at (2.35,11.20) {\textbf{Camera clip}\\Pose, time};

            \draw[draw=black!25, rounded corners=4pt, fill=white, line width=0.6pt] (6.55,10.50) rectangle (18.95,12.40);
            \node[anchor=north west, font=\sffamily\bfseries\footnotesize] at (6.68,12.34) {iw.encode\enspace\textmd{\color{black!55}}};
            \node[model] (embed) at (8.40,11.15) {\textbf{Embed}\\Video--text embedder};
            \node[model] (venc) at (12.80,11.15) {\textbf{Tokenize}\\VLM vision encoder};
            \node[model] (wenc) at (17.20,11.15) {\textbf{Compress}\\Video VAE encoder};
            \draw[flow] (clip.east) -- (6.55,11.20);

            \draw[black!40, dashed, line width=0.5pt] (0.40,9.45) -- (19.15,9.45);
            \node[note] at (3.35,9.45) {only these cross the network};

            \draw[lane] (4.70,6.75) rectangle (19.15,9.00);
            \node[lanetitle] at (4.85,8.90) {Fleet memory};
            \node[memory] (index) at (8.40,7.55) {\textbf{Index}\\Verdict, vector,\\pose, time};
            \node[memory, text width=6.74cm] (store) at (15.275,7.55) {\textbf{Object store}\\Vision tokens, latents};
            \draw[ref] (index.east) -- node[note, pos=0.5] {reference} (store.west);

            \draw[flow] (embed.south) -- node[note, pos=0.46] {retrieval vector} (index.north);
            \draw[flow] (venc.south) -- node[note, pos=0.46] {vision tokens} (venc.south |- store.north);
            \draw[flow] (wenc.south) -- node[note, pos=0.46] {video latent} (wenc.south |- store.north);

            \draw[lane] (0.40,2.30) rectangle (4.30,9.00);
            \node[badge] at (0.80,8.62) {2};
            \node[lanetitle] at (1.08,8.88) {Interpret continuously};

            \node[component] (score) at (2.35,7.55) {\textbf{Prompt scoring}\\Occupied vs.\ clear};
            \draw[flow] ([yshift=0.22cm]index.west) -- node[note, above=1pt, pos=0.38, fill=none] {vector} ([yshift=0.22cm]score.east);
            \draw[flow] ([yshift=-0.22cm]score.east) -- node[note, below=1pt, pos=0.62, fill=none] {verdict} ([yshift=-0.22cm]index.west);

            \node[source, minimum height=2.92cm] (map) at (2.35,4.185) {};
            \foreach \c in {0,1,2,3} {
                \foreach \r in {0,1,2} {
                    \fill[black!9, rounded corners=1pt] (1.085+\c*0.65,4.02+\r*0.47) rectangle ++(0.58,0.40);
                }
            }
            \fill[traffic, rounded corners=1pt] (1.085+2*0.65,4.02+0.47) rectangle ++(0.58,0.40);
            \fill[clear, rounded corners=1pt] (1.085+0*0.65,4.02) rectangle ++(0.58,0.40);
            \fill[clear, rounded corners=1pt] (1.085+3*0.65,4.02) rectangle ++(0.58,0.40);
            \fill[clear, rounded corners=1pt] (1.085+1*0.65,4.02+2*0.47) rectangle ++(0.58,0.40);
            \pic at (1.085+1*0.65+0.29,4.02+2*0.47+0.20) {robot};
            \pic at (1.085+2*0.65+0.29,4.02+0.47+0.20) {robot};
            \pic at (1.085+0*0.65+0.29,4.02+0.20) {robot};
            \pic[rotate=180] at (1.085+3*0.65+0.29,4.02+0.20) {robot};
            \node[align=center, anchor=north] at (2.35,3.88) {\textbf{Live semantic map}\\Updated every 10 s};

            \draw[lane] (4.70,2.30) rectangle (19.15,6.20);
            \node[badge] at (5.10,5.82) {3};
            \node[lanetitle] at (5.38,6.08) {Elaborate on demand};

            \node[component] (search) at (8.40,4.90) {\textbf{Search}\\Meaning, place, time};
            \node[model] (dec) at (12.80,4.90) {\textbf{Describe}\\VLM language decoder};
            \node[model] (wdec) at (17.20,4.90) {\textbf{Inspect scene}\\Video VAE decoder};
            \node[model, text width=7.29cm] (agent) at (10.60,3.10) {\textbf{ReMEmbR agent}\\Tool-calling LLM};

            \draw[bidir] (index.south) -- (search.north);
            \draw[flow] (dec.north |- store.south) -- node[note, pos=0.35] {vision tokens} (dec.north);
            \draw[flow] (wdec.north |- store.south) -- node[note, pos=0.35] {video latent} (wdec.north);

            \draw[flow] ([xshift=-0.25cm]search.south |- agent.north) -- node[note, left=1pt, fill=none] {query} ([xshift=-0.25cm]search.south);
            \draw[flow] ([xshift=0.25cm]search.south) -- node[note, right=1pt, fill=none] {memories} ([xshift=0.25cm]search.south |- agent.north);
            \draw[flow] ([xshift=-0.25cm]dec.south |- agent.north) -- node[note, left=1pt, fill=none] {question} ([xshift=-0.25cm]dec.south);
            \draw[flow] ([xshift=0.25cm]dec.south) -- node[note, right=1pt, fill=none] {answer} ([xshift=0.25cm]dec.south |- agent.north);

            \draw[flow] ($(map.east |- agent.west)+(0,0.22)$) -- node[note, above=1pt, pos=0.64, fill=none] {traffic ahead?} ([yshift=0.22cm]agent.west);
            \draw[flow] ([yshift=-0.22cm]agent.west) -- node[note, below=1pt, pos=0.36, fill=none] {yes / no} ($(map.east |- agent.west)+(0,-0.22)$);

            \node[actor] (fleet) at (2.35,0.95) {\textbf{Fleet manager}\\Planning context};
            \node[actor, text width=5.0cm] (op) at (10.60,0.95) {\textbf{Operator}\\{\itshape ``Was aisle 4 blocked this morning?''}};

            \draw[flow] (map.south) -- node[note, right=1pt, pos=0.62, fill=none] {zone states} (fleet.north);
            \draw[flow] ([xshift=-0.25cm]op.north) -- node[note, left=1pt, pos=0.36, fill=none] {question} ([xshift=-0.25cm]op.north |- agent.south);
            \draw[flow] ([xshift=0.25cm]op.north |- agent.south) -- node[note, right=1pt, pos=0.64, fill=none] {answer} ([xshift=0.25cm]op.north);
            \draw[flow] (wdec.south) |- node[note, pos=0.80, above=1pt, fill=none] {scene visuals} (op.east);

            \draw[flow, dashed] (fleet.west) -- (0.02,0.95) -- (0.02,11.65) -- (0.40,11.65);
            \node[note, rotate=90] at (0.02,5.65) {future route updates};
        \end{tikzpicture}%
    }
    \caption{\industrialRemembr{} moves encoding onto the robots and defers text generation to the server, until a question arrives. Each robot encodes its camera clips on board, and only the resulting embedding, vision tokens and video latent are uploaded to the fleet memory. On the server, incoming embeddings are scored continuously to keep a live semantic traffic map up to date for the fleet manager. When a question arrives, the agent searches the memory and has the retrieved clips described on demand, and an operator can reconstruct any scene for inspection.}
    \label{fig:architecture}
\end{figure}
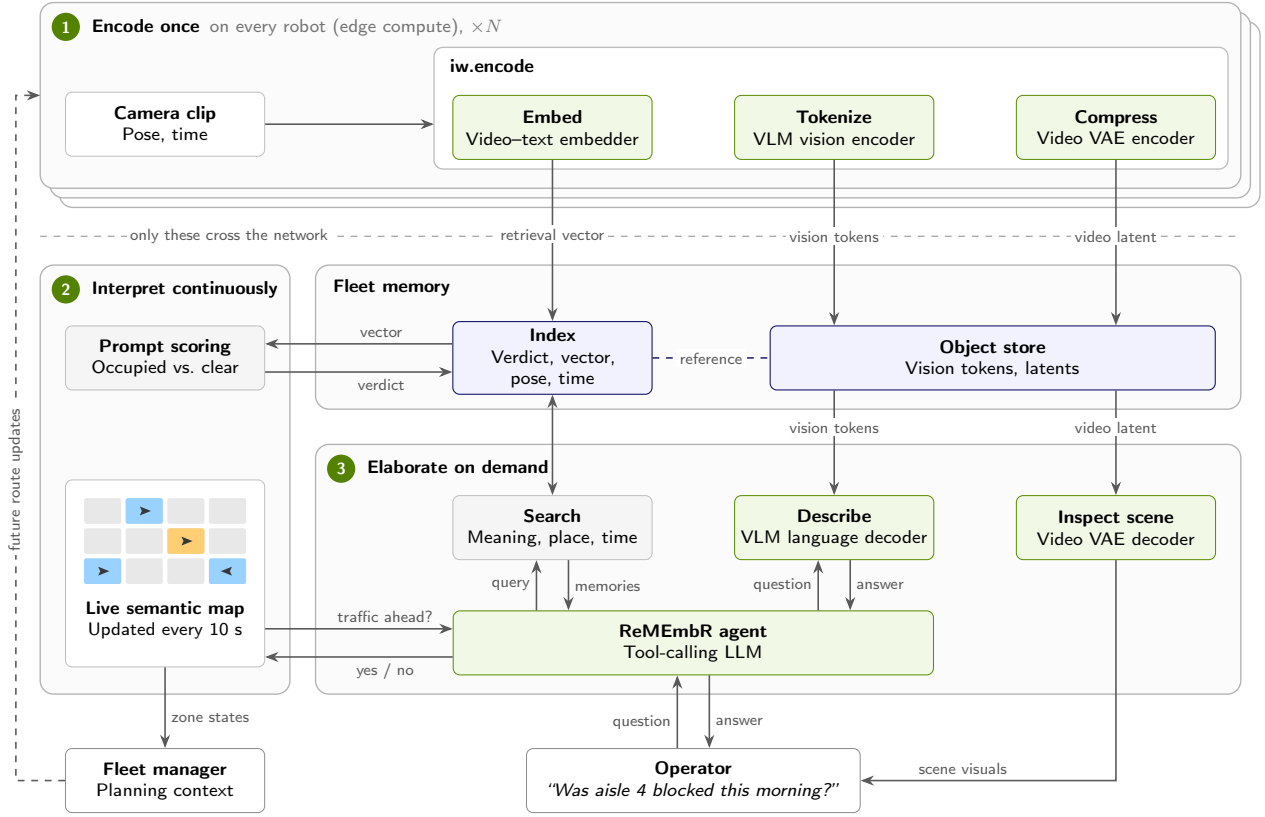

We present \industrialRemembr{} (\FIGREF{fig:architecture}), a shared memory system built around these three principles. It encodes clips on edge compute across the fleet, continuously aggregates and interprets them on server compute for live semantic mapping, and elaborates on any of them on demand. Each robot's onboard Jetson edge-compute device runs \emph{iw.encode}, the edge component for efficient multi-purpose encoding. This deployed service continuously generates video-text embedding vectors for memory retrieval derived from Cosmos-Embed1~\citep{nvidia2025cosmosembed1}, vision tokens for optional scene description derived from the detached vision encoder of Cosmos3-Edge~\citep{nvidia2026cosmos3}, and compressed latent frames for optional scene inspection generated by the Wan2.2 VAE~\citep{wan2025wan,wan2025wan22} encoder. Only the aforementioned multi-purpose scene representations, robot pose and memory timestamp are uploaded into a shared fleet memory, whose vector index is accelerated with NVIDIA cuVS~\citep{rapids2024cuvs}.

To reduce server-side computation, we generate scene descriptions and answers on demand for memories retrieved by the agent, rather than captioning every encoded clip.
We enable this by splitting Cosmos3-Edge at its vision--language boundary, where its server-side language decoder answers questions from cached vision tokens without accessing or re-encoding the original clips.
We deploy a Nemotron-3-Nano-4B~\citep{nvidia2026nemotron3nano,nvidia2025nemotron3} agent through an NVIDIA NIM inference microservice to retrieve memories by content, position, and time, and optionally query the decoder with a text prompt and the retrieved vision tokens.
Operators can also reconstruct scenes from stored latent frames for visual inspection.
On NaVQA, \stackRemembr{}, our version of ReMEmbR updated with the NVIDIA models and a video-based index, achieves comparable retrieval recall while reducing unparseable answers from 4.5\% to 1.9\% with an agent half the parameter size.

We demonstrate the system through a live semantic traffic map that communicates observed aisle conditions to the fleet manager.
We classify each robot's latest observation by comparing its retrieval embedding with paired text prompts describing occupied and clear aisles.
A map service periodically queries the agent for these classifications and updates predefined semantic zones using robot poses already reported to the fleet manager over VDA5050.
In a simulated warehouse, this approach achieves 71.3\% balanced accuracy, exceeding the best caption-based alternative evaluated with the same models by 6.0 percentage points.
We deploy each model as a modular service to support future extensions for station-readiness checks, landmark-based navigation guidance for deployed policy models, and curation of diverse preprocessed training data (Section~\ref{sec:road-ahead}).

Our contributions are:
\begin{itemize}
    \item \textbf{Video-native agentic memory.} We rebuild ReMEmbR on an efficient NVIDIA model stack and move its index key from the caption to the video itself. Retrieval is no longer bottlenecked by what a captioner chose to mention, and a smaller, tool-calling agent returns more reliable structured answers (Section~\ref{sec:nvidia-stack}).
    \item \textbf{Encode once, distributed across the fleet.} Every robot acts as a distributed sensor, and a modular set of parallel encoder services on its Jetson turns each clip, in a single pass, into multi-purpose representations for memory retrieval, description and reconstruction. Per-clip compute is offloaded to the edge and scales out with the fleet, and raw camera frames never leave the robot (Section~\ref{sec:encode}).
    \item \textbf{Disaggregated vision-language inference for describe-on-demand.} Cosmos3-Edge is split at its vision--language boundary, with the vision encoder on the robot and the language decoder on the server. Cached vision tokens are encoded once and queried many times, so description cost scales with the questions asked rather than with the video recorded (Sections~\ref{sec:memory} and~\ref{sec:agent}).
    \item \textbf{A zero-shot, multi-purpose live semantic map.} The retrieval embedding the robot already computes doubles as a training-free semantic classifier, in this case reused for traffic classification, interpreted continuously and fed to the fleet manager over its existing VDA5050 interface. An agent sits between the map and the memory as a natural-language interface, so a new site-level insight requires only a new question, with no new model to train (Section~\ref{sec:traffic}).
    \item \textbf{A modular, scale-out service architecture.} Encoders, a GPU-accelerated vector index, a tiered object store, NIM-served agents and on-demand decoders run as independent, versioned services that can each be scaled or upgraded in isolation. We profile the edge encoders on Jetson and model how server compute, bandwidth and storage scale from a single robot to a full fleet (Sections~\ref{sec:placement} and~\ref{sec:results-fleet}).
\end{itemize}
\section{Preliminaries}
\label{sec:starting-point}

\industrialRemembr builds on three main lines of research: large language model (LLM) agents that support versatile tool use, retrieval-augmented generation from semantic embeddings, and the ReMEmbR system, which combines the two to encode and query a robot's memories over long horizons.

\textbf{Agents.} In this context, an agent is an LLM that runs in an interactive loop with access to predefined tools. At each step in the loop, the model reads the task and its own running history, and then either calls any of the available tools or provides its final answer. ReAct~\citep{yao2023react} features the model writing out its intermediate reasoning between tool calls, so that the result of one call guides the next. Practically, every tool call must be well-formatted to allow for deterministic parsing and automated execution. The loop also needs a limit on the allowed number of steps, so that it always ends with a concrete answer. \industrialRemembr follows this loop, and adds a tool that lets the agent ask a vision--language decoder its own question about any memory it retrieves.

\textbf{Retrieval-augmented generation.} Retrieval-augmented generation (RAG)~\citep{lewis2020rag} allows a model to derive answers from a corpus even if it is too large to directly fit in context. The corpus is chunked and indexed in advance, and for each question the $k$ most relevant chunks are retrieved and incorporated in the model's context window. Retrieval is usually done with a \emph{bi-encoder}, which embeds queries and documents separately so that search involves a fast vector comparison~\citep{karpukhin2020dpr}. The trade-off is that every document vector is computed before any query is known. A \emph{cross-encoder}, by contrast, reads a query and a document together, which tends to be more accurate but often too slow to apply to a whole corpus, so it normally only re-ranks a shortlist already retrieved~\citep{nogueira2019passage} by the bi-encoder. \emph{Agentic RAG}~\citep{singh2025agenticrag} exposes retrieval as a tool for the agent. The agent writes its own retrieval queries, retries upon search misses, and decides when it has sufficient evidence to return with a contextually grounded response. \industrialRemembr uses agentic RAG over video instead of text, with a video--text bi-encoder as the index and a decoder that reads the question and the retrieved memory together.

\textbf{ReMEmbR.} ReMEmbR~\citep{anwar2025remembr} lets a user ask a mobile robot where something is, when something happened, or what it saw over hours of operation, and returns answers the robot can act on, such as an $(x,y,z)$ position to navigate toward. Because a question rarely needs the whole history $H$, ReMEmbR answers from a small retrieved subset $R \subseteq H$, approximating $p(A \mid H, Q)$ with $p(A \mid R, Q)$ for a question $Q$ and answer $A$.

It works in two phases. In the \emph{memory-building} phase, VILA~\citep{lin2024vila} captions each 3\,s video segment, MxBai~\citep{lee2024mxbai} embeds the caption, and Milvus~\citep{wang2021milvus} stores the embedding with the segment's mean position and time.\footnote{Position and orientation are stored in separate columns.} In the \emph{querying} phase, an LLM agent searches this memory by text, position or time for up to three rounds, and then returns a structured JSON answer that holds text, a yes/no value, a time, a duration or a goal position. The accompanying NaVQA benchmark contains 210 questions over seven CODa sequences~\citep{zhang2024coda} of 15 to 30 minutes each. A position answer counts as correct within 15\,m, and a time answer within 2\,min.

The memory is built before any question is asked, so it has to be general enough to serve whichever question comes later. However, ReMEmbR captions every clip when it is recorded. The memory can answer only what the captioner chose to mention, and generating a caption for every clip is the main cost of writing a memory.

\stackRemembr keeps its pipeline and updates its models (Section~\ref{sec:nvidia-stack}). \industrialRemembr keeps the two-phase design but stores no caption, and describes a memory only once a question asks about it (Section~\ref{sec:redesign}).
\section{\stackRemembr{}: Updated Models and Video-Based Retrieval}
\label{sec:nvidia-stack}

In ReMEmbR, a caption does two jobs. It serves as the \emph{index key}, since its text embedding decides which questions can find the clip, and it serves as the \emph{context}, since it is all the agent reads once the memory is retrieved. Both jobs depend on what the captioner chose to mention when first encoding. ``A forklift in a warehouse'' may be accurate but still leave out whether the forklift blocks the aisle, in which case a question about blocked aisles becomes less likely to align with the caption's embedding. Captioning is therefore an information bottleneck for retrieval, since a scene detail has to survive both the captioner's limited attention and the embedding step to play any role in the search. To build toward \industrialRemembr{}, we first introduce \stackRemembr{}. It keeps ReMEmbR's pipeline, updates its underlying models, and moves the index key from the caption to the video itself, circumventing the retrieval bottleneck.

\stackRemembr{} retains ReMEmbR's three-second segments, its position and time metadata, and its search by content, position and time, so that any change on NaVQA can be attributed to the models. Two of the three model updates are direct replacements. Cosmos3-Edge~\citep{nvidia2026cosmos3} takes over captioning from VILA-1.5-13B~\citep{lin2024vila}. It is roughly one-third the size, and its emphasis on physical-world and spatial understanding suits questions about how a scene is arranged. Nemotron-3-Nano-4B~\citep{nvidia2026nemotron3nano,nvidia2025nemotron3} takes over the agent role from Llama-3.1-8B~\citep{grattafiori2024llama3}, ReMEmbR's locally hosted agent. It is a hybrid Mamba-Transformer model with half the parameters, post-trained for tool use and structured output, and it has access to the same search tools as before.

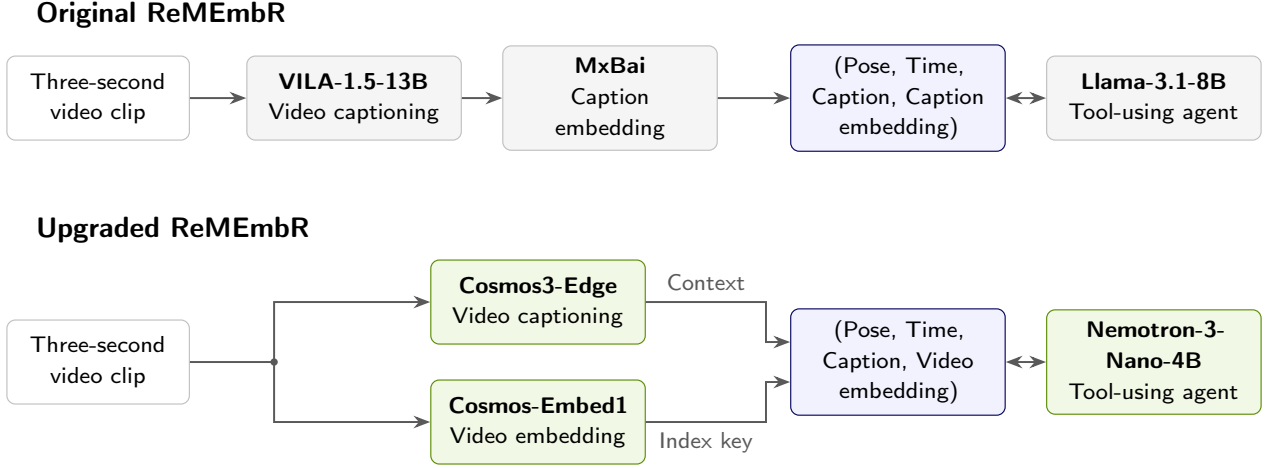
\begin{figure}[H]
    \centering
    \resizebox{\linewidth}{!}{%
        \begin{tikzpicture}[
            font=\sffamily\footnotesize,
            component/.style={draw=black!25, fill=black!4, rounded corners=3pt, text width=2.4cm, minimum height=1.05cm, align=center, inner sep=4pt},
            replacement/.style={component, draw=nvidiagreen!75!black, fill=nvidiagreen!10},
            memory/.style={component, draw=blue!35!black, fill=blue!5},
            clip/.style={component, text width=2.0cm, fill=white},
            flow/.style={-{Stealth[length=2.1mm]}, draw=black!65, line width=0.7pt},
            note/.style={font=\sffamily\footnotesize, text=black!65, fill=white, inner sep=2pt}
        ]
            \node[anchor=west, font=\sffamily\bfseries] at (-0.1,4.35) {Original ReMEmbR};
            \node[clip] (original-video) at (0.8,3.3) {Three-second video clip};
            \node[component] (original-caption) at (4.0,3.3) {\textbf{VILA-1.5-13B}\\Video captioning};
            \node[component] (original-embed) at (7.2,3.3) {\textbf{MxBai}\\Caption\\embedding};
            \node[memory] (original-memory) at (10.8,3.3) {(Pose, Time, Caption, Caption embedding)};
            \node[component] (original-agent) at (14.0,3.3) {\textbf{Llama-3.1-8B}\\Tool-using agent};
            \draw[flow] (original-video) -- (original-caption);
            \draw[flow] (original-caption) -- (original-embed);
            \draw[flow] (original-embed) -- (original-memory);
            \draw[flow, {Stealth[length=2.1mm]}-{Stealth[length=2.1mm]}] (original-memory) -- (original-agent);

            \node[anchor=west, font=\sffamily\bfseries] at (-0.1,1.65) {\stackRemembr};
            \node[clip] (rebuilt-video) at (0.8,0) {Three-second video clip};
            \node[replacement] (rebuilt-caption) at (6.3,0.75) {\textbf{Cosmos3-Edge}\\Video captioning};
            \node[replacement] (rebuilt-embed) at (6.3,-0.75) {\textbf{Cosmos-Embed1}\\Video embedding};
            \node[memory] (rebuilt-memory) at (10.8,0) {(Pose, Time, Caption, Video embedding)};
            \node[replacement] (rebuilt-agent) at (14.0,0) {\textbf{Nemotron-3-}\\\textbf{Nano-4B}\\Tool-using agent};
            \coordinate (rebuilt-input) at (3.0,0);
            \draw[draw=black!65, line width=0.7pt] (rebuilt-video.east) -- (rebuilt-input);
            \fill[black!65] (rebuilt-input) circle (1.3pt);
            \draw[flow] (rebuilt-input) |- (rebuilt-caption.west);
            \draw[flow] (rebuilt-input) |- (rebuilt-embed.west);
            \draw[flow] (rebuilt-caption.east) -- (9.15,0.75) -- (9.15,0.25) -- ([yshift=0.25cm]rebuilt-memory.west);
            \draw[flow] (rebuilt-embed.east) -- (9.15,-0.75) -- (9.15,-0.25) -- ([yshift=-0.25cm]rebuilt-memory.west);
            \node[note] at (8.4,1.0) {Context};
            \node[note] at (8.4,-1.0) {Index key};
            \draw[flow, {Stealth[length=2.1mm]}-{Stealth[length=2.1mm]}] (rebuilt-memory) -- (rebuilt-agent);
        \end{tikzpicture}%
    }
    \caption{\stackRemembr indexes video directly instead of caption text, so retrieval is no longer limited to what the captioner chose to say. Original ReMEmbR is shown on top and \stackRemembr at the bottom, with the replaced components in green. In the original, the index is built from caption text, so retrieval can find only what the captioner chose to say. \stackRemembr sends each clip down two independent paths: Cosmos-Embed1 produces the video embedding that serves as the index key, and Cosmos3-Edge produces the caption the agent reads once an observation is retrieved. Both workflows still caption every segment in advance. Bidirectional arrows denote agent tool calls and returned evidence.}
    \label{fig:nvidia-rebuild-workflow}
\end{figure}

The third update changes what the index is built from. ReMEmbR embeds the caption text with MxBai~\citep{lee2024mxbai}. \stackRemembr{} embeds the clip itself with Cosmos-Embed1~\citep{nvidia2025cosmosembed1}, which maps video and text into a shared semantic space, so a natural-language query is matched against the content of the video (\FIGREF{fig:nvidia-rebuild-workflow}). The caption still accompanies each retrieved memory as context, but it no longer limits what the search can find.

\stackRemembr{} leaves the caption in place as context, so generating it remains the bulk of the cost of writing a memory. Section~\ref{sec:results} evaluates each of the three replacements on NaVQA, changing one component at a time.

\section{\industrialRemembr{}: Encode Once, Interpret Continuously, Elaborate on Demand}
\label{sec:redesign}

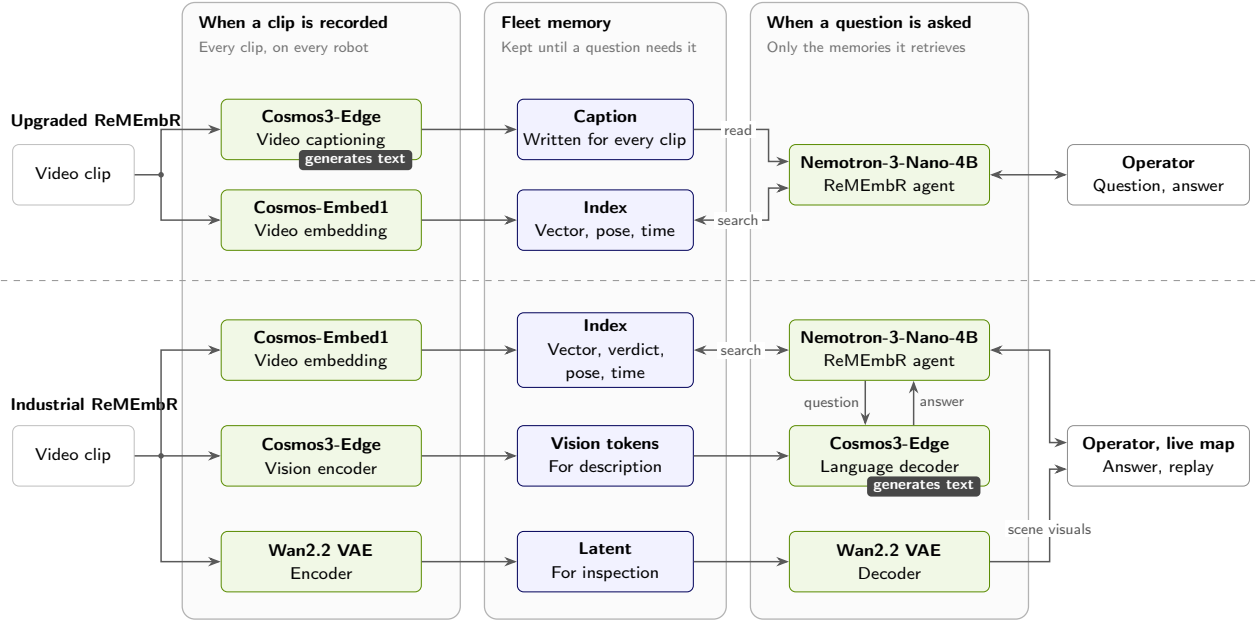
\begin{figure}[H]
    \centering
    \resizebox{\linewidth}{!}{%
        \begin{tikzpicture}[
            font=\sffamily\footnotesize,
            lane/.style={draw=black!30, rounded corners=6pt, fill=black!1.5, line width=0.6pt},
            lanetitle/.style={anchor=north west, font=\sffamily\bfseries\footnotesize},
            lanesub/.style={anchor=north west, font=\sffamily\scriptsize, text=black!55},
            rowtitle/.style={anchor=south west, font=\sffamily\bfseries\footnotesize, align=left, inner sep=0pt},
            component/.style={draw=black!25, fill=black!4, rounded corners=3pt, text width=3.1cm, minimum height=1.0cm, align=center, inner sep=3pt},
            model/.style={component, draw=nvidiagreen!75!black, fill=nvidiagreen!10},
            memory/.style={component, draw=blue!35!black, fill=blue!5, text width=2.7cm},
            actor/.style={component, fill=white, draw=black!45, text width=2.8cm},
            clip/.style={component, text width=1.8cm, fill=white},
            flow/.style={-{Stealth[length=2.1mm]}, draw=black!65, line width=0.7pt},
            bidir/.style={{Stealth[length=2.1mm]}-{Stealth[length=2.1mm]}, draw=black!65, line width=0.7pt},
            note/.style={font=\sffamily\scriptsize, text=black!65, fill=white, inner sep=1.5pt},
            tag/.style={rounded corners=2pt, fill=black!72, text=white, font=\sffamily\bfseries\scriptsize, inner xsep=3pt, inner ysep=2pt}
        ]
            \draw[lane] (3.10,0.25) rectangle (7.70,10.45);
            \node[lanetitle] at (3.25,10.35) {When a clip is recorded};
            \node[lanesub] at (3.25,9.93) {Every clip, on every robot};

            \draw[lane] (8.10,0.25) rectangle (12.10,10.45);
            \node[lanetitle] at (8.25,10.35) {Fleet memory};
            \node[lanesub] at (8.25,9.93) {Kept until a question needs it};

            \draw[lane] (12.50,0.25) rectangle (17.10,10.45);
            \node[lanetitle] at (12.65,10.35) {When a question is asked};
            \node[lanesub] at (12.65,9.93) {Only the memories it retrieves};

            \draw[black!40, dashed, line width=0.5pt] (0.10,5.85) -- (20.85,5.85);

            \node[rowtitle] at (0.25,8.35) {\stackRemembr};
            \node[clip] (clip1) at (1.30,7.60) {Video clip};

            \node[model] (capt) at (5.40,8.35) {\textbf{Cosmos3-Edge}\\Video captioning};
            \node[model] (embed1) at (5.40,6.85) {\textbf{Cosmos-Embed1}\\Video embedding};
            \node[tag, anchor=east] at ([xshift=-0.15cm]capt.south east) {generates text};

            \node[memory] (cap) at (10.10,8.35) {\textbf{Caption}\\Written for every clip};
            \node[memory] (index1) at (10.10,6.85) {\textbf{Index}\\Vector, pose, time};

            \node[model] (agent1) at (14.80,7.60) {\textbf{Nemotron-3-Nano-4B}\\ReMEmbR agent};
            \node[actor] (op1) at (19.25,7.60) {\textbf{Operator}\\Question, answer};

            \coordinate (fan1) at (2.75,7.60);
            \draw[draw=black!65, line width=0.7pt] (clip1.east) -- (fan1);
            \fill[black!65] (fan1) circle (1.3pt);
            \draw[flow] (fan1) |- (capt.west);
            \draw[flow] (fan1) |- (embed1.west);
            \draw[flow] (capt.east) -- (cap.west);
            \draw[flow] (embed1.east) -- (index1.west);
            \draw[flow] (cap.east) -- (12.80,8.35) |- ([yshift=0.22cm]agent1.west);
            \draw[bidir] (index1.east) -- (12.80,6.85) |- ([yshift=-0.22cm]agent1.west);
            \node[note] at (12.30,8.35) {read};
            \node[note] at (12.30,6.85) {search};
            \draw[bidir] (agent1.east) -- (op1.west);

            \node[rowtitle] at (0.25,3.70) {\industrialRemembr};
            \node[clip] (clip2) at (1.30,2.95) {Video clip};

            \node[model] (embed2) at (5.40,4.70) {\textbf{Cosmos-Embed1}\\Video embedding};
            \node[model] (venc) at (5.40,2.95) {\textbf{Cosmos3-Edge}\\Vision encoder};
            \node[model] (wenc) at (5.40,1.20) {\textbf{Wan2.2 VAE}\\Encoder};

            \node[memory] (index2) at (10.10,4.70) {\textbf{Index}\\Vector, verdict,\\pose, time};
            \node[memory] (tok) at (10.10,2.95) {\textbf{Vision tokens}\\For description};
            \node[memory] (lat) at (10.10,1.20) {\textbf{Latent}\\For inspection};

            \node[model] (agent2) at (14.80,4.70) {\textbf{Nemotron-3-Nano-4B}\\ReMEmbR agent};
            \node[model] (dec) at (14.80,2.95) {\textbf{Cosmos3-Edge}\\Language decoder};
            \node[model] (wdec) at (14.80,1.20) {\textbf{Wan2.2 VAE}\\Decoder};
            \node[tag, anchor=east] at ([xshift=-0.15cm]dec.south east) {generates text};

            \node[actor] (op2) at (19.25,2.95) {\textbf{Operator, live map}\\Answer, replay};

            \coordinate (fan2) at (2.75,2.95);
            \draw[draw=black!65, line width=0.7pt] (clip2.east) -- (fan2);
            \fill[black!65] (fan2) circle (1.3pt);
            \draw[flow] (fan2) |- (embed2.west);
            \draw[flow] (fan2) -- (venc.west);
            \draw[flow] (fan2) |- (wenc.west);
            \draw[flow] (embed2.east) -- (index2.west);
            \draw[flow] (venc.east) -- (tok.west);
            \draw[flow] (wenc.east) -- (lat.west);

            \draw[bidir] (index2.east) -- node[note] {search} (agent2.west);
            \draw[flow] (tok.east) -- (dec.west);
            \draw[flow] (lat.east) -- (wdec.west);
            \draw[flow] ([xshift=-0.40cm]agent2.south) -- node[note, left=1pt, fill=none] {question} ([xshift=-0.40cm]dec.north);
            \draw[flow] ([xshift=0.40cm]dec.north) -- node[note, right=1pt, fill=none] {answer} ([xshift=0.40cm]agent2.south);

            \draw[bidir] (agent2.east) -- (17.45,4.70) |- ([yshift=0.22cm]op2.west);
            \draw[flow] (wdec.east) -- (17.45,1.20) |- ([yshift=-0.22cm]op2.west);
            \node[note] at (17.45,1.75) {scene visuals};
        \end{tikzpicture}%
    }
    \caption{\industrialRemembr defers text generation until a question arrives, whereas \stackRemembr (top) fixes the text an answer can draw on when the clip is recorded. \stackRemembr runs the full Cosmos3-Edge captioner on every clip; \industrialRemembr (bottom) runs only encoders at that point and keeps their outputs in fleet memory. The language half of Cosmos3-Edge runs once a question has arrived, on the memories the agent retrieves, and the Wan2.2 decoder turns a stored latent back into video when someone wants to see the scene. The dark tag marks the step that generates text. Green boxes are models, blue boxes are storage.}
    \label{fig:encode-once}
\end{figure}

\stackRemembr{} showed that a memory can be found without its caption, yet it still generates one for every clip before any question is known. \industrialRemembr{} goes further and eliminates that step. Each robot stores the inputs that a server-side model would need to describe a memory, and the description itself is generated only when a question calls for one (\FIGREF{fig:encode-once}).

\label{sec:principles}
Three principles guide the design. The first is to \emph{encode once}. Each clip passes through the robot's encoders a single time and comes out as the representations that later steps read: an embedding for retrieval, vision tokens for description, and a video latent for inspection. No later step has to return to the camera frames, which never need to leave the robot. Per-clip work therefore runs on the robot, and its cost grows with the length of the footage. The second is to \emph{interpret continuously}. The server scores each incoming embedding as it arrives, which keeps the live semantic map current without generating any text. The third is to \emph{elaborate on demand}. Text is generated once a question is known, so it can address that question, and the same memory can be described again for the next one. Per-question work therefore runs on the server, and its cost grows with the number of questions asked rather than with the amount of video recorded. The following subsections follow these principles in turn: \emph{encode once} on the robot (Section~\ref{sec:encode}), \emph{interpret continuously} on the memory the robots feed (Section~\ref{sec:memory}), and \emph{elaborate on demand} on the server (Section~\ref{sec:agent}).

\subsection{On the Robot: One Pass, Three Representations}
\label{sec:encode}

Each robot runs iw.encode, a set of three encoder services on its Jetson AGX Xavier, on every three-second segment. Each encoder produces the input for a later application, and none of them needs the camera frames again.

Cosmos-Embed1 maps the segment to a single 768-dimensional vector in a shared video--text space, which takes 3~kB. This vector is the memory's index key, and the agent's content search reads it (Section~\ref{sec:agent}). As in \stackRemembr{}, it replaces ReMEmbR's MxBai embedding of the caption, so retrieval no longer depends on what a caption mentions. Text prompts map into the same space, so the server can score the same vector for the traffic verdict of Section~\ref{sec:traffic} and can support other prompt-based checks, each at the cost of a dot product.

The vision encoder of Cosmos3-Edge produces 60 projected vision tokens of width 2048, which the model's language decoder would normally receive from it directly. The robot stores these tokens with the metadata the decoder needs to read them, 247~kB in total. The language decoder on the server reads them when the agent asks about a retrieved memory (Section~\ref{sec:agent}). They replace the caption that ReMEmbR generates for every clip, so the robot never runs the expensive step of generating text on edge.

The Wan2.2 VAE~\citep{wan2025wan,wan2025wan22} encoder compresses the segment into a $48 \times 2 \times 30 \times 52$ video latent of 300~kB, one-twentieth of the size of its uncompressed frames. The Wan2.2 decoder on the server reads this latent when an operator wants to revisit a memory, and reconstructs the scene for inspection. ReMEmbR has no counterpart to it, since its memory keeps the caption's embedding and not a compression of the scene itself.

The three encoders run in parallel as independent services, and their outputs are the only data that leave the robot (\FIGREF{fig:architecture}). Each memory covers three seconds in our configuration, and how often a robot writes one is a deployment setting. NaVQA uses contiguous segments; fleet resource projections assume one write every 12~s, while the live map polls every 10~s. Encoder timing and memory on Xavier are reported in \TABREF{tab:xavier-latency}, \TABREF{tab:encoder-memory} and Appendix~\ref{app:xavier}.

\subsection{Fleet Memory}
\label{sec:memory}

Fleet memory has two parts, matched to how each kind of data is used. Every segment adds one row to a shared Milvus collection~\citep{wang2021milvus}, containing its Cosmos-Embed1 vector, the robot's pose, the capture time and any verdict the server computes as the row arrives, which is what lets the server \emph{interpret continuously} without reading the larger data. Indexes over the vector, the position and the time support search by content, place and time, and NVIDIA cuVS~\citep{rapids2024cuvs} builds and searches the vector index on the GPU. The vision tokens and the latent are much larger and are read only when a memory is retrieved, so they are kept in an object store, and the row holds a reference to them.

ReMEmbR's reference deployment runs Milvus on the robot~\citep{anwar2025remembr}. Here, we centralize it on the on-premises server to offload database and index-serving work from robots, avoid duplicated fleet indexes and per-robot reconciliation, and provide the fleet a shared live database of memories to independently contribute to and query from.

\subsection{On the Server: Answering Questions}
\label{sec:agent}
\label{sec:placement}

Questions are answered on an on-premises server by a Nemotron-3-Nano-4B agent. As in ReMEmbR, the agent searches the index by content, position and time, and the entries it retrieves accumulate in its working memory. Each retrieved entry now holds a reference to its cached vision tokens in place of a caption, and a fourth tool lets the agent put a question of its own to the Cosmos3-Edge language decoder about any entry. The agent therefore decides both which moments to look at and what to ask about them.

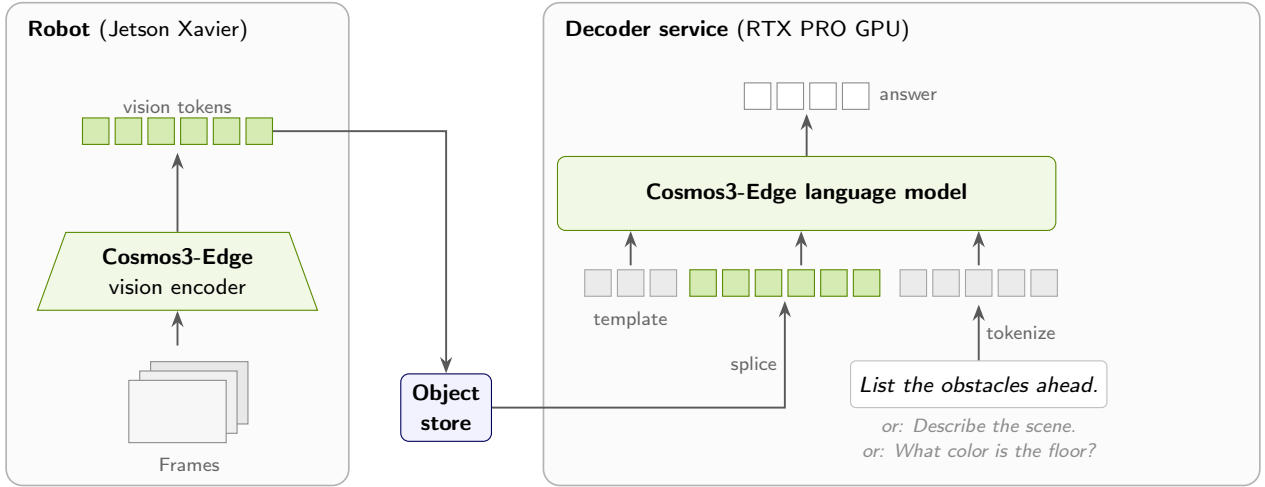
\begin{figure}[H]
    \centering
    \resizebox{\linewidth}{!}{%
        \begin{tikzpicture}[
            font=\sffamily\footnotesize,
            lane/.style={draw=black!30, rounded corners=6pt, fill=black!1.5, line width=0.6pt},
            lanetitle/.style={anchor=north west, font=\sffamily\bfseries\footnotesize},
            vtok/.style={draw=nvidiagreen!75!black, fill=nvidiagreen!30, minimum size=3.4mm, inner sep=0pt},
            ttok/.style={draw=black!35, fill=black!8, minimum size=3.4mm, inner sep=0pt},
            otok/.style={draw=black!45, fill=white, minimum size=3.4mm, inner sep=0pt},
            model/.style={draw=nvidiagreen!75!black, fill=nvidiagreen!10, rounded corners=3pt, align=center, inner sep=3pt},
            enc/.style={trapezium, trapezium angle=70, shape border rotate=0, draw=nvidiagreen!75!black, fill=nvidiagreen!10, align=center, minimum width=2.9cm, minimum height=1.0cm, inner sep=2pt},
            memory/.style={draw=blue!35!black, fill=blue!5, rounded corners=3pt, align=center, inner sep=4pt},
            frame/.style={draw=black!45, fill=white, minimum width=1.25cm, minimum height=0.8cm, inner sep=0pt},
            txt/.style={draw=black!25, fill=white, rounded corners=2pt, minimum height=6mm, inner sep=3pt},
            flow/.style={-{Stealth[length=2.1mm]}, draw=black!65, line width=0.7pt},
            note/.style={font=\sffamily\scriptsize, text=black!60, inner sep=1.5pt}
        ]
            \draw[lane] (0,0) rectangle (4.4,6.2);
            \node[lanetitle] at (0.15,6.1) {Robot \textmd{(Jetson Xavier)}};

            \foreach \k in {2,1,0} {\node[frame, fill=black!\the\numexpr 3+\k*3\relax] (f\k) at (2.2+\k*0.14,0.95+\k*0.12) {};}
            \node[note, anchor=north] at (2.35,0.42) {Frames};

            \node[enc] (venc) at (2.2,2.75) {\textbf{Cosmos3-Edge}\\vision encoder};
            \draw[flow] ([yshift=0.2cm]f2.north -| venc.south) -- (venc.south);

            \foreach \i in {0,...,5} {\node[vtok] (r\i) at (1.15+\i*0.42,4.55) {};}
            \draw[flow] (venc.north) -- ([yshift=-1mm]$(r2.south)!0.5!(r3.south)$);
            \node[note, anchor=south] at ($(r2.north)!0.5!(r3.north)$) {vision tokens};

            \node[memory] (store) at (5.65,1.0) {\textbf{Object}\\\textbf{store}};
            \draw[flow] (r5.east) -- (5.65,4.55) -- (store.north);

            \draw[lane] (6.9,0) rectangle (16.1,6.2);
            \node[lanetitle] at (7.05,6.1) {Decoder service \textmd{(RTX PRO GPU)}};

            \foreach \i in {0,1,2} {\node[ttok] (t\i) at (7.6+\i*0.42,2.6) {};}
            \foreach \i in {0,...,5} {\node[vtok] (v\i) at (8.95+\i*0.42,2.6) {};}
            \foreach \i in {0,...,4} {\node[ttok] (q\i) at (11.65+\i*0.42,2.6) {};}

            \node[note] at (8.02,2.12) {template};

            \draw[flow] (store.east) -- (10.0,1.0) -- node[note, left=1pt, pos=0.4, align=right] {splice} (10.0,2.37);

            \node[txt] (qtext) at (12.49,1.3) {\itshape List the obstacles ahead.};
            \draw[flow] (qtext.north) -- node[note, right=1pt] {tokenize} ([yshift=-1mm]q2.south);
            \node[note, anchor=north, align=center, text=black!45] at (12.49,0.92) {\itshape or: Describe the scene.\\\itshape or: What color is the floor?};

            \node[model, minimum width=6.4cm, minimum height=0.95cm] (llm) at (10.28,3.75) {\textbf{Cosmos3-Edge language model}};
            \foreach \n in {t1,v3,q2} {\draw[flow] ([yshift=0.5mm]\n.north) -- (\n.north |- llm.south);}

            \foreach \i in {0,...,3} {\node[otok] (o\i) at (9.65+\i*0.42,5.0) {};}
            \draw[flow] (llm.north -| o1.south) ++(0.21,0) -- ([yshift=-0.5mm]$(o1.south)!0.5!(o2.south)$);
            \node[note, anchor=west] at ([xshift=2pt]o3.east) {answer};
        \end{tikzpicture}%
    }
    \caption{Cached vision tokens let the server describe a scene on demand, without the source frames. The robot runs only the vision half of Cosmos3-Edge and stores its vision tokens (green). When a question arrives, the server's language model receives the usual chat template and the embedded question (gray), with the cached vision tokens spliced into the slot where encoded video would normally go, after checking that the token counts match. The same cached tokens can answer any number of different questions.}
    \label{fig:describe-on-demand}
\end{figure}

This is where the server \emph{elaborates on demand}. The decoder answers from the cached tokens alone (\FIGREF{fig:describe-on-demand}). It builds its usual prompt from the chat template and the question, and it places the stored vision tokens in the slot where freshly encoded video would normally go. This split turns an intermediate model output into a lasting memory. Edge--server partitioning such as Neurosurgeon~\citep{kang2017neurosurgeon} chooses a split layer per request to reduce its latency or energy, and the intermediate output is discarded once that request is served. Here, the boundary sits at the model's vision--language interface, and the robot's tokens persist in fleet memory long after they are encoded. One set of tokens can answer any number of questions.

After one or more rounds of search, the agent returns a structured answer with text, yes/no, position, time and duration fields that downstream systems can consume directly. An operator can query the agent interactively, and a service can call it on a schedule, as the live map does in Section~\ref{sec:traffic}. When an operator wants to see a scene, the Wan2.2 decoder reconstructs it from the stored latent. New typed tools can expose site-specific tasks to the agent without any change to how memories are encoded.

Every component runs as a versioned service on the local network. Our reference setup places fleet memory, cuVS, both decoders and the NIM-served agent on one on-premises server with NVIDIA RTX PRO 6000 Blackwell Server Edition GPUs. Larger sites can distribute the agent and the decoders across dedicated GPUs and provision Milvus, the object store and the map service separately, and each modular service can be scaled or upgraded in isolation (Appendix~\ref{app:agent}).
\section{A First Application: The Live Semantic Traffic Map}
\label{sec:traffic}

We return to the blocked aisle example from the introduction. The first robot to pass the clutter stores a memory of it like any other clip, and whether the aisle ahead is occupied is already implicit in that memory's embedding. When the live semantic map next refreshes, it finds that this embedding vector aligns more closely with those of descriptions of an occupied aisle than those of a clear one, and the aisle's zone turns orange. The fleet manager can then take the blocked aisle into account when planning routes in that area. Further, an operator who wants to know what exactly is in the way can click the robot and read a description of what it saw, generated on the spot from the memory's cached vision tokens. None of this changes the robots, which run the same iw.encode pipeline as in Section~\ref{sec:encode}. The server adds two cached text prototypes and a map service that queries the agent on a fixed clock, and the rest is \industrialRemembr{} (\FIGREF{fig:online-agentic-map}).

\label{sec:reporting}
The report adds no robot-side computation, as it reuses the Cosmos-Embed1 vector $\mathbf{v}$ generated for retrieval as part of the deployed iw.encode pipeline. Cosmos-Embed1 projects video and text onto a shared embedding space, so a written description can be directly compared with a recorded video. We describe each of the two outcomes with a handful of concise text prompts, such as ``a warehouse aisle blocked by a pallet'' for an occupied aisle and ``an empty warehouse aisle'' for a clear one (Appendix~\ref{app:prompts}). Averaging the embeddings of each set and rescaling the result to unit length gives one \emph{prototype} per outcome, $\mathbf{p}_{\mathrm{occ}}$ and $\mathbf{p}_{\mathrm{clr}}$, following the prompt ensembling that CLIP introduced for zero-shot classification~\citep{radford2021clip}. The server caches both prototypes, so the prompt ensembles can be maintained and updated. Unlike CLIP's evaluation, where each image is embedded solely to be classified, the verdict reuses the vector the robot already uploads as the memory's index key, so it requires no separate classification pass. As each vector arrives, the server's \emph{verdict} on the clip is \emph{traffic} when the clip is closer to the occupied prototype,
\begin{equation}
    \mathbf{v}^{\top}\mathbf{p}_{\mathrm{occ}} \;>\; \mathbf{v}^{\top}\mathbf{p}_{\mathrm{clr}},
    \label{eq:verdict}
\end{equation}
and \emph{clear} otherwise. The traffic verdict is thus derived on the server straight from the embedding the robot already computes at encoding time, at the cost of two dot products, without a detector, captioner or any other trained model dedicated to traffic classification.

\input{figures/fig_semantic_map}

A single similarity score is a weak signal. A clip's embedding describes the whole scene, including its background, colors and lighting, so its similarity to any short prompt is low, and the scores for the two outcomes end up close together. This score attenuation is a direct consequence of the average pooling applied to the Q-Former tokens in the Cosmos-Embed1 architecture. Because this pooling operation collapses all visual tokens into a single global vector, v encodes the entire scene thereby diluting the specific signals. By contrasting v against the two opposing, generalized prototypes, we isolate the relevant semantic axis and bypass the dilution caused by the global visual pooling. A fixed threshold on either score would be fragile, so the verdict compares the two directly. Their difference, $\mathbf{v}^{\top}(\mathbf{p}_{\mathrm{occ}} - \mathbf{p}_{\mathrm{clr}})$, cancels whatever the prototypes have in common, such as the aisle itself, and keeps only how much more the clip looks like an occupied aisle than a clear one.

The server stores the verdict in the same memory row as the embedding, its pose, its timestamp and the reference to the clip's cached vision tokens. Next to the verdict, the row holds a short sentence such as ``In the aisle straight ahead there is traffic,'' filled in from a fixed template. The agent can read this sentence directly, and writing the memory still generates no text.

\label{sec:mapping}
\label{sec:map-to-routing}
The map combines this memory with what the fleet manager already knows. Under VDA5050, each AMR regularly sends the fleet manager a \texttt{state} message whose \texttt{agvPosition} field reports the robot's pose, its position $(x, y)$ and heading $\theta$. Fleet memory adds what each robot sees. The map service divides the floor plan into \emph{zones}, polygons that each cover an aisle segment or a crossing, and keeps one traffic state per zone.

\begin{figure}[!t]
\centering
\definecolor{lmTraffic}{RGB}{253,203,110}
\definecolor{lmTrafficInk}{RGB}{122,78,0}
\definecolor{lmClear}{RGB}{157,209,251}
\definecolor{lmIdle}{RGB}{232,232,232}
\definecolor{lmLane}{RGB}{250,250,250}
\definecolor{lmLaneEdge}{RGB}{189,189,189}
\definecolor{lmInk}{RGB}{51,51,51}
\definecolor{lmMuted}{RGB}{110,110,110}
\definecolor{lmGreen}{RGB}{84,138,20}
\definecolor{lmGreenFill}{RGB}{238,246,227}
\definecolor{lmBlue}{RGB}{26,35,126}
\definecolor{lmBlueFill}{RGB}{238,240,255}
\resizebox{\linewidth}{!}{%
\begin{tikzpicture}[
    x=1cm, y=1cm,
    font=\sffamily\footnotesize, text=lmInk,
    >={Stealth[length=1.6mm, width=1.3mm]},
    lm-lane/.style={draw=lmLaneEdge, fill=lmLane, rounded corners=3pt, line width=0.5pt},
    lm-tile/.style={rounded corners=1.5pt, minimum width=1.9cm, minimum height=1.25cm,
                    inner sep=0pt, anchor=south west},
    lm-id/.style={font=\sffamily\scriptsize, text=lmMuted, anchor=north west, inner sep=2.5pt},
    lm-state/.style={font=\sffamily\scriptsize\bfseries, anchor=south west, inner sep=2.5pt},
    lm-tag/.style={font=\sffamily\scriptsize, text=lmMuted, inner sep=1pt},
    lm-card/.style={draw=lmLaneEdge, fill=white, rounded corners=2pt, line width=0.4pt},
    lm-key/.style={anchor=west, text=lmMuted, inner sep=0pt},
    lm-val/.style={anchor=west, inner sep=0pt},
    lm-btn/.style={draw=lmLaneEdge, fill=white, rounded corners=5pt, line width=0.4pt,
                   minimum height=0.44cm, minimum width=2.1cm, inner sep=0pt},
    lm-model/.style={draw=lmGreen, fill=lmGreenFill, rounded corners=2pt, line width=0.5pt,
                     minimum height=0.85cm, minimum width=2.3cm, align=center, inner sep=2pt},
    lm-store/.style={draw=lmBlue, fill=lmBlueFill, rounded corners=2pt, line width=0.5pt,
                     minimum height=0.85cm, minimum width=1.75cm, align=center, inner sep=2pt},
    lm-flow/.style={->, draw=lmInk, line width=0.45pt},
    lm-robot/.pic={\fill[lmInk] (-0.15,-0.105) -- (0.15,0) -- (-0.15,0.105) -- (-0.07,0) -- cycle;},
    lm-memory/.pic={\draw[lmBlue, fill=white, line width=0.7pt] (0,0) circle[radius=0.075];},
  ]

  \draw[lm-lane] (0,0)    rectangle (8.65,5.75);
  \draw[lm-lane] (8.95,0) rectangle (15.00,5.75);
  \node[anchor=west, inner sep=0pt] at (0.30,5.42)
    {\textbf{Live semantic map}\enspace{\scriptsize\color{lmMuted}each robot asked every 10\,s}};
  \node[anchor=west, inner sep=0pt] at (9.25,5.42)
    {\textbf{Selected robot}\enspace{\scriptsize\color{lmMuted}on click}};

  \foreach \row/\y in {A/3.75, B/2.35, C/0.95} {
    \foreach \col/\x in {1/0.30, 2/2.35, 3/4.40, 4/6.45} {
      \node[lm-tile, fill=lmIdle] (z\row\col) at (\x,\y) {};
    }
  }
  \node[lm-tile, fill=lmClear]   at (zB1.south west) {};
  \node[lm-tile, fill=lmClear]   at (zB2.south west) {};
  \node[lm-tile, fill=lmClear]   at (zC3.south west) {};
  \node[lm-tile, fill=lmTraffic] at (zA4.south west) {};
  \foreach \row in {A, B, C} {
    \foreach \col in {1, 2, 3, 4} {
      \node[lm-id] at (z\row\col.north west) {\row\col};
    }
  }
  \node[lm-state] at (zB1.south west) {clear};
  \node[lm-state] at (zB2.south west) {clear};
  \node[lm-state] at (zC3.south west) {clear};
  \node[lm-state, text=lmTrafficInk] at (zA4.south west) {traffic};

  \foreach \z/\a/\lab/\d in {zB1/180/R1/-1, zC3/0/R3/1, zA4/0/R2/1} {
    \coordinate (\lab) at ($(\z.center)+(\d*0.22,-0.08)$);
    \pic[rotate=\a] at (\lab) {lm-robot};
    \pic at ($(\lab)+(\d*-0.44,0)$) {lm-memory};
    \node[lm-tag, anchor=south] at ($(\z.center)+(0,0.23)$) {\lab};
  }
  \coordinate (sel) at (R2);
  \draw[lmInk, line width=0.6pt] (sel) circle[radius=0.27];

  \begin{scope}[shift={(0.30,0.42)}]
    \fill[lmTraffic, rounded corners=1pt] (0.00,-0.08) rectangle +(0.26,0.16);
    \node[lm-tag, anchor=west] at (0.32,0) {traffic};
    \fill[lmClear, rounded corners=1pt]   (1.25,-0.08) rectangle +(0.26,0.16);
    \node[lm-tag, anchor=west] at (1.57,0) {clear};
    \fill[lmIdle, rounded corners=1pt]    (2.35,-0.08) rectangle +(0.26,0.16);
    \node[lm-tag, anchor=west] at (2.67,0) {not asked};
    \pic at (4.25,0) {lm-robot};
    \node[lm-tag, anchor=west] at (4.44,0) {robot pose};
    \pic at (5.95,0) {lm-memory};
    \node[lm-tag, anchor=west] at (6.14,0) {retrieved memory};
  \end{scope}

  \draw[lm-card] (9.25,3.05) rectangle (14.70,5.10);
  \node[lm-val] at (9.45,4.78) {\textbf{Robot 02}\enspace{\color{lmMuted}zone A4}};
  \draw[lmLaneEdge, line width=0.3pt] (9.45,4.52) -- (14.50,4.52);
  \node[lm-key] at (9.45,4.22) {Agent answer};
  \node[fill=lmTraffic, text=lmTrafficInk, rounded corners=4pt, inner xsep=4pt, inner ysep=1.5pt,
        font=\sffamily\scriptsize\bfseries, anchor=west] (ans) at (12.05,4.22) {yes};
  \node[lm-val] at ($(ans.east)+(0.12,0)$) {traffic};
  \node[lm-key] at (9.45,3.77) {Robot pose};
  \pic at (12.20,3.77) {lm-robot};
  \node[lm-val] at (12.47,3.77) {VDA5050 state};
  \node[lm-key] at (9.45,3.32) {Retrieved memory};
  \pic at (12.20,3.32) {lm-memory};
  \node[lm-val] at (12.47,3.32) {pose, time};
  \draw[lmInk, line width=0.45pt] ($(sel)+(0.27,0)$) -- (9.25,0 |- sel);

  \node[anchor=west, inner sep=0pt] at (9.25,2.68)
    {\textbf{Describe on request}\enspace{\scriptsize\color{lmMuted}from cached tokens}};
  \node[lm-btn] (b1) at (10.55,2.12) {Traffic status};
  \node[lm-btn] (b2) at (13.40,2.12) {Full scene};
  \node[lm-store] (store) at (10.125,0.80) {\textbf{Object store}\\[-1pt]{\scriptsize vision tokens}};
  \node[lm-model] (dec)   at (13.55,0.80) {\textbf{Cosmos3-Edge}\\[-1pt]{\scriptsize Language decoder}};
  \draw[lm-flow] (store) -- node[above, lm-tag, yshift=1pt] {tokens} (dec);
  \draw[lm-flow] (b2.south) -- node[right, lm-tag, xshift=1pt] {prompt} (b2.south |- dec.north);
  \draw[lm-flow] (b1.south) -- ++(0,-0.30) -| ($(dec.north)+(-0.60,0)$);
\end{tikzpicture}
}
\caption{Zone colors encode the agent's latest answer per zone, and an operator can request a description of the memory behind it. Each zone takes the color of the agent's latest answer for the robot inside it, and each robot is drawn at its VDA5050 pose beside the memory the agent retrieved for it. Selecting a robot shows its answer and lets the operator request a description, which the Cosmos3-Edge decoder writes from the vision tokens already in memory. The zone layout and robots are illustrative.}
\label{fig:live-map}
\end{figure}
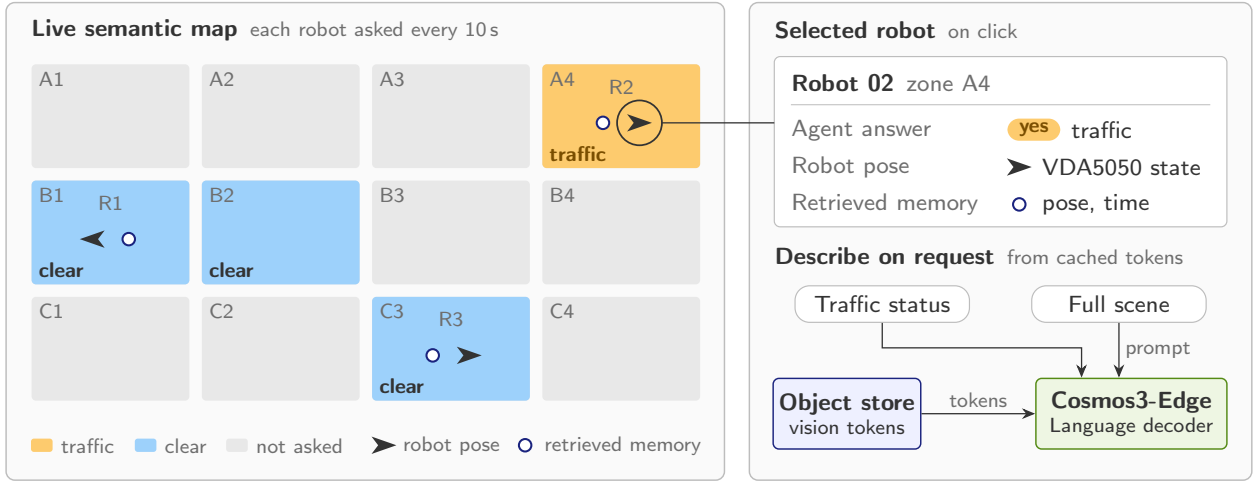

To minimize redundancy, the map service does not react to every memory a robot writes. It polls on a fixed period of $\Delta t$ seconds (10~s by default). The request rate follows the robot count and the poll interval. The compute each request takes also depends on the agent's work and the length of its answer. On each tick it takes the latest pose of every robot that reported since the last one and looks up which zone the robot is in. Robots outside all zones are skipped. For the rest, it asks the agent a question that carries the robot's time and position:
\begin{quote}
\itshape ``The time now is HH:MM:SS. You are a robot at position $[x, y, z]$ in a warehouse. \ldots\ Is there any other vehicle or agent in the zone directly ahead of you right now? Answer with only one word yes or no.'' (full text in Appendix~\ref{app:prompts-map})
\end{quote}
The agent answers it the way it answers an operator, by searching memory with its text, position and time tools. Each query runs with a fresh agent state, so no memories carry over from the previous robot and the tool-call budget is reset. A yes labels the zone as \emph{traffic} and a no relabels it as \emph{clear}. Any other output triggers a re-query (\FIGREF{fig:live-map}).

%

While updating the map does not strictly need the agent in this narrow case, since the verdict is already available with each memory, we maintain the agent as an adaptive natural language interface. It decides how to answer, relying on a robot's latest memory or checking it against what other robots recorded in the same aisle shortly before. The map service only sends it a question, and a different question entails no further changes across the pipeline. We see traffic as the first layer of a multi-purpose semantic map of industrial sites, whose zones could eventually record whether a charging station is free, whether a dock is filling up, or whether a recognizable landmark is still where a navigation policy expects it. Building a dedicated classifier for each of these would mean new models, new memory fields and new robot software every time a unique need arose. With an agent exposing a natural language interface between the semantic map and the shared memory, each new need becomes a new question posed to the same interface.

\begin{figure}[t]
    \centering
    \includegraphics[width=0.8\textwidth]{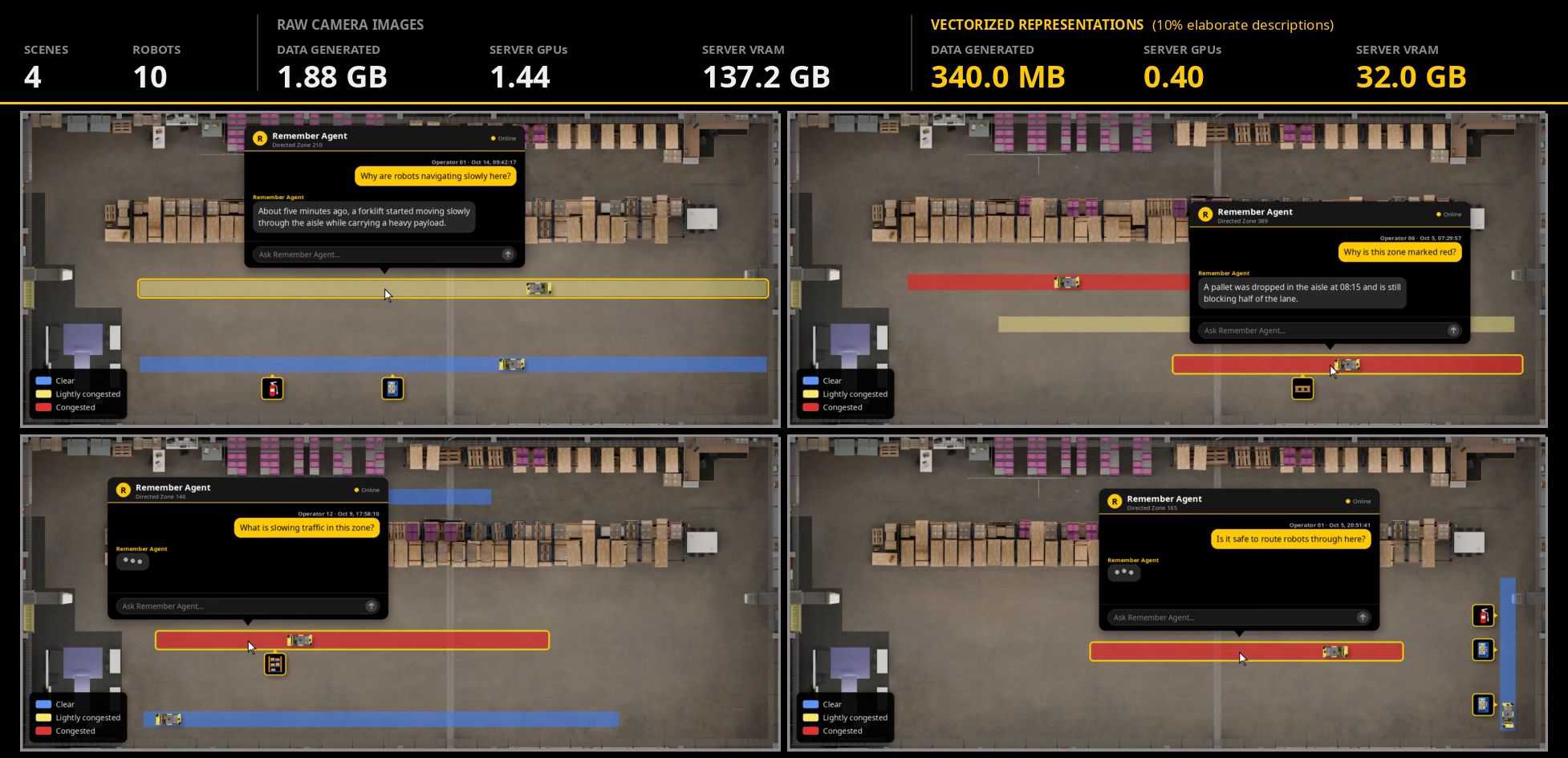}
    \caption{Operators can request an on-demand description from the \industrialRemembr{} frontend. The popover window allows users to generate on-demand descriptions using vision tokens.}
    \label{fig:semantic-frontend}
\end{figure}

Each update is pushed to a web page as soon as it is computed. Robot markers follow the reported poses, and zone colors follow the latest answers. Beside each robot, the map also marks where the memory behind its answer was recorded, so an operator can see what the answer was based on. Clicking a robot asks for more detail about its latest memory (\FIGREF{fig:live-map}, right; \FIGREF{fig:semantic-frontend}). The Cosmos3-Edge language decoder reads that memory's cached vision tokens together with a general request to describe the scene, and the operator receives a conditionally generated description.

\section{Results}
\label{sec:results}

Sections~\ref{sec:nvidia-stack}--\ref{sec:traffic} make three changes to ReMEmbR. They replace its models and index the video itself, they store encoder outputs in place of captions, and they read a live traffic map from those outputs. Our experiments test whether each change does what it was meant to do, and whether the resulting system remains affordable across a fleet. They are organized around four questions:
\begin{itemize}
    \item Do the new models retrieve and answer better? (Section~\ref{sec:results-navqa})
    \item Can one encoding answer any later question? (Section~\ref{sec:results-describing})
    \item Does the cost scale to a fleet? (Section~\ref{sec:results-fleet})
    \item Can the memory keep a live traffic map? (Section~\ref{sec:results-traffic})
\end{itemize}
\TABREF{tab:glance} summarizes the main results.

\begin{table}[b]
\centering\footnotesize
\begin{tabular}{@{}llrrl@{}}
\toprule
Result & Compared & Baseline & This work & Section \\
\midrule
First-attempt parse failures (NaVQA) & Original vs.\ \stackRemembr{} & 20.7\% & \textbf{13.8\%} & \ref{sec:results-navqa} \\
Unscorable answers (NaVQA) & Original vs.\ \stackRemembr{} & 4.5\% & \textbf{1.9\%} & \ref{sec:results-navqa} \\
Retrieval recall (NaVQA) & Caption-text vs.\ video index & 20.8\% & \textbf{24.6\%} & \ref{sec:results-navqa} \\
Description GPU equiv. (100 robots) & Every clip vs.\ 10\% described & 36.25 & \textbf{1.04} & \ref{sec:results-fleet} \\
Traffic balanced accuracy & Best caption vs.\ embedding & 65.3\% & \textbf{71.3\%} & \ref{sec:results-traffic} \\
\bottomrule
\end{tabular}
\caption{Main results and their baselines.}
\label{tab:glance}
\end{table}

\subsection{Do the New Models Retrieve and Answer Better?}
\label{sec:results-navqa}
\label{sec:results-finding}
\label{sec:results-endtoend}

\stackRemembr{} replaces all three models of ReMEmbR. Nemotron-3-Nano-4B takes over from Llama-3.1-8B as the agent, Cosmos3-Edge from VILA-1.5-13B as the captioner, and a Cosmos-Embed1 video index from the MxBai caption-text index. The pipeline around these models is unchanged, so any difference on NaVQA comes from the models, and swapping in one model at a time shows which model causes which difference. NaVQA~\citep{anwar2025remembr} tests long-horizon memory over all seven CODa campus sequences, where each memory holds one three-second clip with its capture pose and time. We run three trials per configuration, each answering 67 yes/no questions, for 201 responses (Appendix~\ref{app:navqa}). \emph{Retrieval recall} is the fraction of annotated evidence segments that appear in any of the agent's tool results, so it depends on both the index and the queries the agent writes. Reliability is measured by \emph{first-attempt parse failures}, answers the evaluator cannot parse on the first try, and by \emph{unscorable answers}, which stay unparsable after retries. Position errors are medians over parsed answers.

\begin{table}[t]
    \centering
    \footnotesize
    \setlength{\tabcolsep}{6pt}
    \begin{tabular}{@{}llcccc@{}}
        \toprule
        System & Swapped in & First-try & Unscorable & Recall & Position \\
               &            & fail.\ (\%) $\downarrow$ & (\%) $\downarrow$ & (\%) $\uparrow$ & err.\ (m) $\downarrow$ \\
        \midrule
        Original ReMEmbR & --                       & 20.7          & 4.5          & 22.9          & 69.1 \\
        \midrule
        One swap         & Agent: Nemotron          & \textbf{16.2} & \textbf{0.9} & 18.7          & 71.0 \\
                         & Captioner: Cosmos3-Edge  & 22.4          & 3.6          & 26.7          & 74.6 \\
                         & Index: Cosmos-Embed1     & 23.5          & 5.5          & \textbf{29.7} & 79.4 \\
        \midrule
        \stackRemembr{}  & All three                & 13.8          & 1.9          & 24.6          & 66.6 \\
        \bottomrule
    \end{tabular}
    \caption{Swapping one component of original ReMEmbR at a time on full NaVQA, three trials per configuration. Original ReMEmbR uses Llama-3.1-8B, VILA-1.5-13B and MxBai. Each one-swap row replaces one of them and keeps the other two. Bold marks the best single swap in each column where it beats original ReMEmbR. \TABREF{tab:grid} reports all eight configurations.}
    \label{tab:swap}
\end{table}

Each new model improves a different measure (\TABREF{tab:swap}). Nemotron is the one that makes answers reliable. Swapping it in alone cuts first-attempt parse failures from 20.7\% to 16.2\% and unscorable answers from 4.5\% to 0.9\%, whereas swapping in the captioner or the index instead raises first-attempt failures to 22.4\% and 23.5\%. The effect does not depend on what the agent reads. Across all four captioner--index pairs of \TABREF{tab:grid}, Nemotron stays within 12.2--16.2\% first-attempt failures against 20.7--23.5\% for Llama, and within 0.9--1.9\% unscorable answers against 3.6--5.5\%. Nemotron has half the parameters of Llama, so this gain likely reflects its post-training for tool use and structured output rather than its size.

Recall, in contrast, follows the index. Replacing the caption-text index with Cosmos-Embed1 alone raises recall from 22.9\% to 29.7\%, the largest gain of any single swap, and the video index matches or improves recall for every agent--captioner pair. With Nemotron as the agent, it adds 8.5 points over caption-text indexing with VILA captions and 3.8 points with Cosmos3-Edge captions. This is the bottleneck of Section~\ref{sec:nvidia-stack}. A caption-text index can find only what the captioner chose to mention, while the video index matches a question against the clip itself, and so it needs neither a caption nor a caption-text embedding stage. \industrialRemembr{} relies on this property when it stops writing captions altogether.

Combined, the three models improve on original ReMEmbR in both reliability and recall. First-attempt parse failures fall from 20.7\% to 13.8\% and unscorable answers from 4.5\% to 1.9\%, relative reductions of about 33\% and 58\%, and median position error falls from 69.1 to 66.6~m. Recall rises from 22.9\% to 24.6\%, less than the index swap alone achieves. Because recall also depends on the agent's queries, Nemotron lowers it by 2.3--5.9 points relative to Llama in every pair, and the full stack keeps only part of the index's gain. In exchange, Nemotron lowers median position error by 7.7--11.9~m relative to Llama whenever it is paired with either new component.

\stackRemembr{} therefore finds more evidence and answers more reliably, yet it still captions every clip as it is written. The next question is whether \industrialRemembr{} can drop that step without losing what the caption provided.

\subsection{Can One Encoding Answer Any Later Question?}
\label{sec:results-describing}

\industrialRemembr{} stores Cosmos3-Edge vision tokens in place of a caption and splits the model at its vision--language boundary (Section~\ref{sec:redesign}). The split is only worthwhile if decoding from stored tokens gives the same answer as running the full model on the frame, and if elaborating a memory is fast enough to fit inside an agent's answer. We test both on a separate run of 150 traffic frames (Appendix~\ref{app:describe}). The vision encoder encodes each frame once, and the language decoder answers three questions from the cached tokens. The same frames and questions also run through the unsplit model for comparison. All timings use one NVIDIA RTX PRO 6000 Blackwell Server Edition GPU.

\begin{table}[H]
    \centering
    \footnotesize
    \setlength{\tabcolsep}{10pt}
    \begin{tabular}{@{}lllrr@{}}
        \toprule
        Operation     & Reads from memory     & Output            & Tokens & Time (s) \\
        \midrule
        Describe      & Vision tokens, 247~kB & One-word answer   & 2      & 0.03 \\
                      &                       & List of obstacles & 43     & 0.38 \\
                      &                       & Scene description & 146    & 1.25 \\
        \midrule
        Inspect scene & Latent, 300~kB        & Five-frame scene  &        & 1.76 \\
        \bottomrule
    \end{tabular}
    \caption{Time to elaborate one memory on the server. Descriptions use the Cosmos3-Edge language decoder and scene inspection the Wan2.2 decoder. Description times and token counts are means over the 150 frames of that run; the inspection time is a median.}
    \label{tab:ondemand}
\end{table}

The split model reproduces the full model exactly. Under greedy decoding, decoding from cached tokens produced the same text as the unsplit Cosmos3-Edge for all three questions on every frame, 450 of 450 cases, and tokens written to the object store and read back later decoded through the same path. Splitting the model therefore leaves its outputs unchanged, and a later question can be answered from memory as if the frame were still at hand.

\input{figures/fig_one_cache_many_questions}

Encoding is paid once per memory and decoding once per question. Encoding a frame's vision tokens takes 0.056~s, after which the same tokens answer a one-word check, a list of obstacles or a full scene description (\FIGREF{fig:one-cache}). Under these fixed input and decoding settings, a 146-token description takes 1.25~s (\TABREF{tab:ondemand}); generation time is approximately proportional to output length for this workload.

\begin{figure}[t]
    \centering
    \sffamily\footnotesize
    \setlength{\tabcolsep}{1.5pt}
    \renewcommand{\arraystretch}{0.6}
    \newlength{\reconw}\setlength{\reconw}{0.315\linewidth}
    \newcommand{\reconimg}[1]{\raisebox{-0.5\height}{#1}}
    \newcommand{\reconrow}[1]{\rotatebox[origin=c]{90}{#1}}
    \begin{tabular}{@{}c@{\hspace{4pt}}ccc@{}}
                                  & Forklift ahead           & Robot ahead           & Clear aisle \\[4pt]
        \reconrow{Reference}   & \reconimg{\includegraphics[width=\reconw]{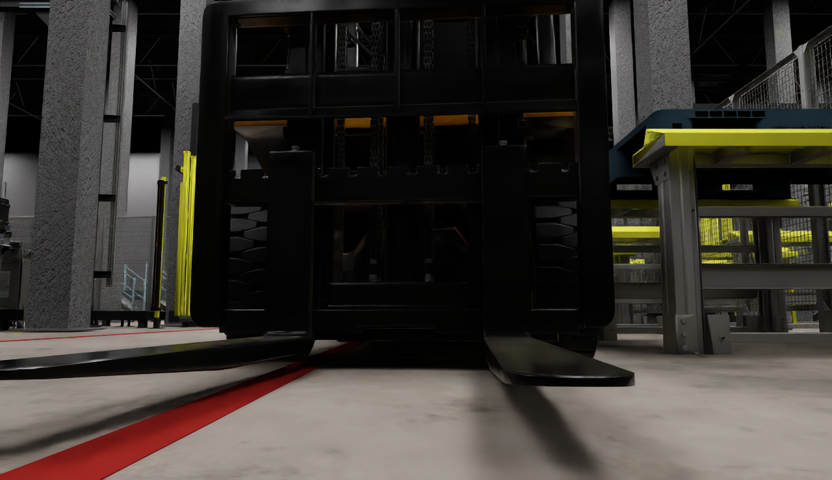}}  & \reconimg{\includegraphics[width=\reconw]{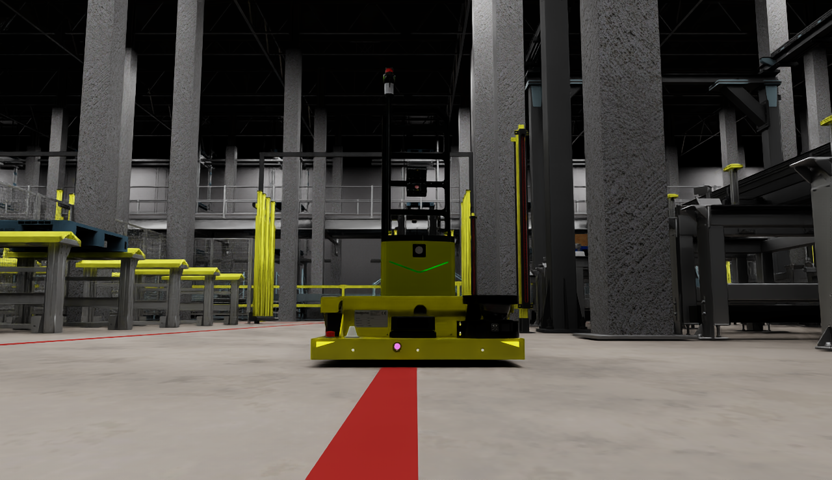}}  & \reconimg{\includegraphics[width=\reconw]{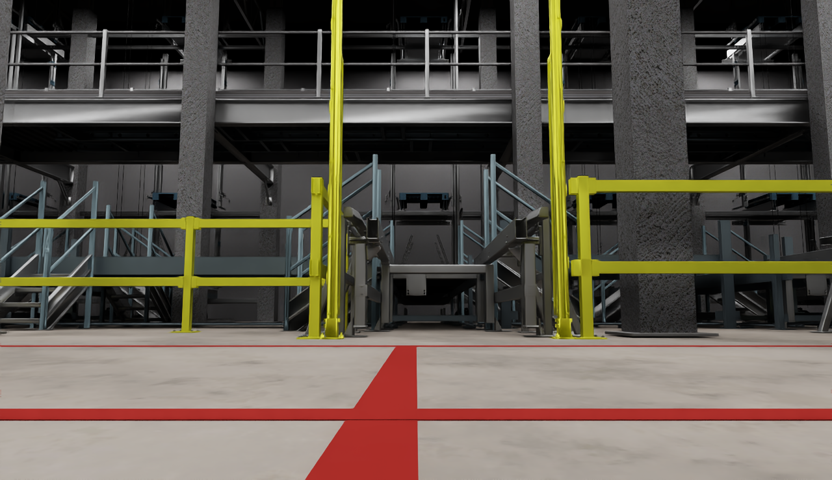}} \\[1.05cm]
        \reconrow{Reproduced}  & \reconimg{\includegraphics[width=\reconw]{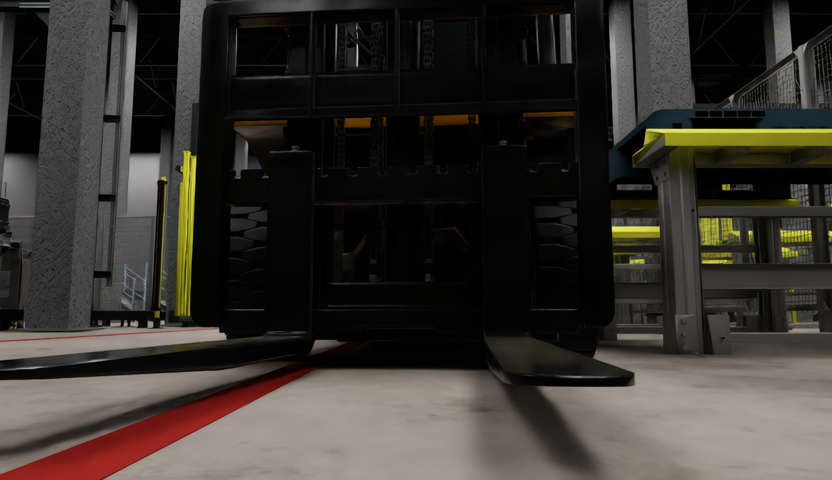}} & \reconimg{\includegraphics[width=\reconw]{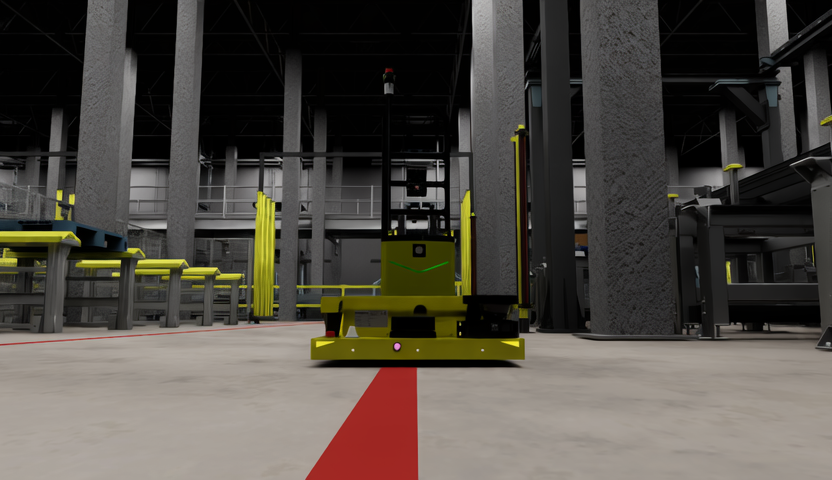}} & \reconimg{\includegraphics[width=\reconw]{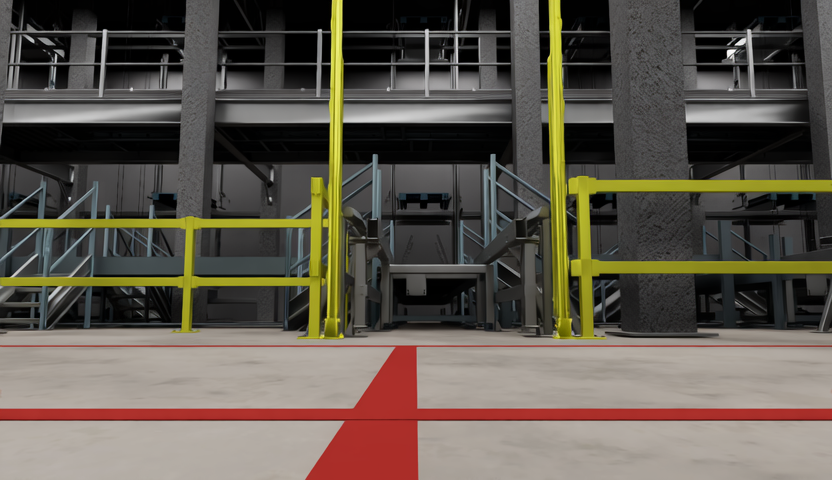}} \\[1.05cm]
        \reconrow{Error}       & \reconimg{\includegraphics[width=\reconw]{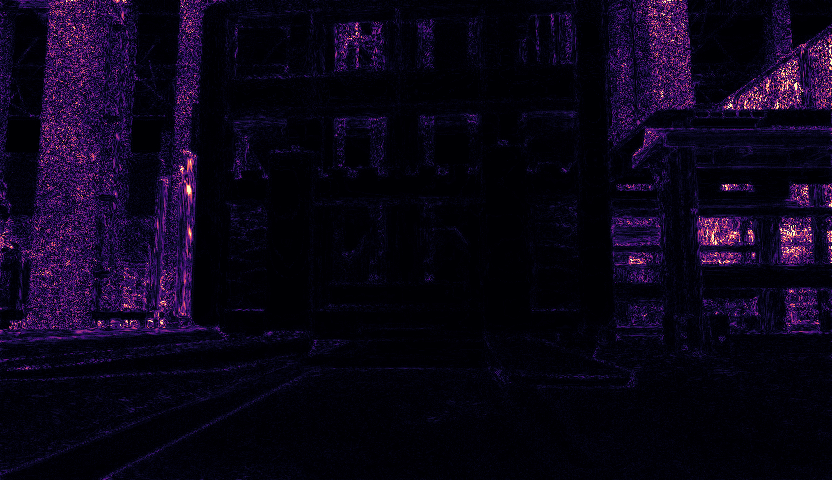}}   & \reconimg{\includegraphics[width=\reconw]{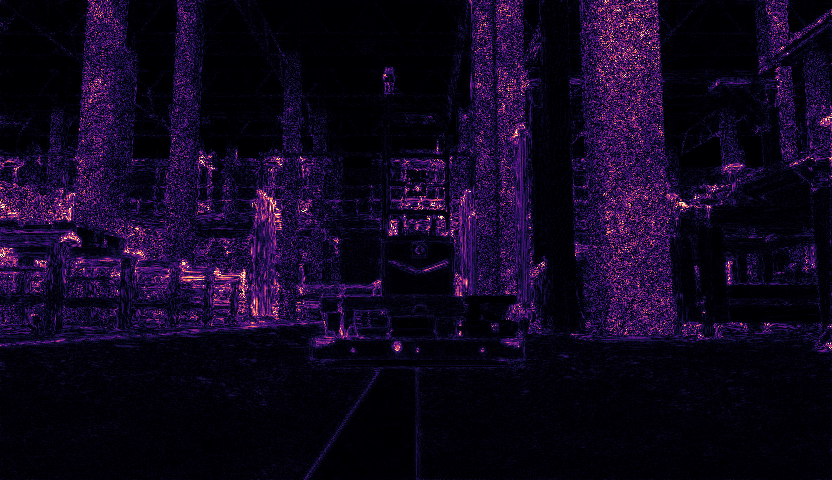}}   & \reconimg{\includegraphics[width=\reconw]{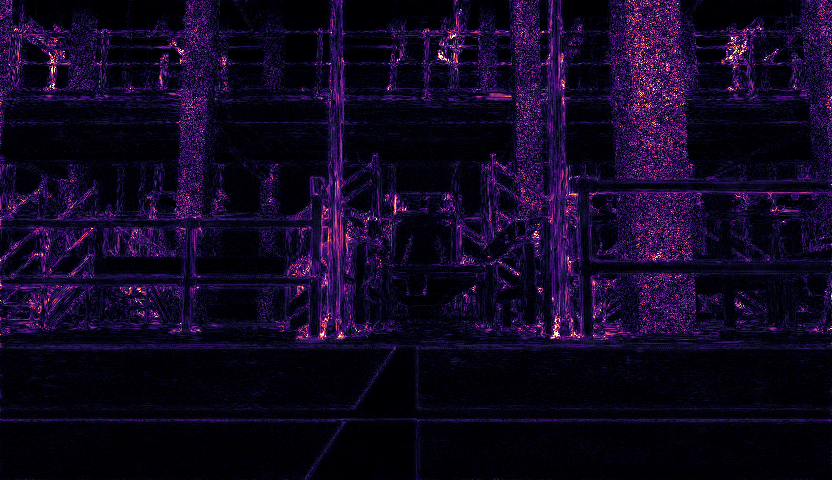}} \\
    \end{tabular}
    \caption{A stored latent is enough to reproduce a scene on the server for inspection. Shown are three observations of the traffic run. \textbf{Top:}~reference observations used in the fidelity evaluation. \textbf{Middle:}~the scene reconstructed on the server from each memory's stored latent. \textbf{Bottom:}~absolute pixel differences, with full brightness marking an error of at least 10\% of the intensity range. The examples cover a forklift ahead, a robot ahead and a clear aisle and were selected for scene content variety.}
    \label{fig:reconstruction}
\end{figure}

Reconstructing a scene for human inspection works the same way. A five-frame 480$\times$832 clip is stored as a $48 \times 2 \times 30 \times 52$ BF16 latent of 300~kB, one-twentieth of its uncompressed size; casting the FP32 encoder output to BF16 halves storage and changes the reconstruction metrics below by under 0.1\% (\TABREF{tab:bf16}), and the Wan2.2 decoder has a median reconstruction time of 1.76~s (\TABREF{tab:ondemand}). Reconstructions of the 150 frames reach 39.7~dB PSNR, 0.974 SSIM~\citep{wang2004ssim} and 0.0052 LPIPS~\citep{zhang2018lpips}, with the remaining differences mostly in fine textures and thin edges (\FIGREF{fig:reconstruction}). An operator who inspects a memory therefore sees a close reconstruction of the frames the robot recorded.

\subsection{Does the Cost Scale to a Fleet?}
\label{sec:results-fleet}
\label{sec:results-cost}
\label{sec:results-scale}

\industrialRemembr{} moves encoding onto the robots and generation onto the questions, so its cost has an edge side and a server side, which we measure separately. Edge timings use warmed TensorRT~8.5.2 engines on a Jetson AGX Xavier with the inputs of \TABREF{tab:encoders}, and they include transfers between host and device (Appendix~\ref{app:xavier}). Server measurements use one RTX PRO 6000 Blackwell Server Edition GPU unless stated otherwise. Where we extrapolate to a fleet, we assume one memory per robot every 12~s over an eight-hour day and mark the result as modeled (Appendix~\ref{app:fleet}). The live map's 10~s poll interval is a separate setting (Section~\ref{sec:traffic}).

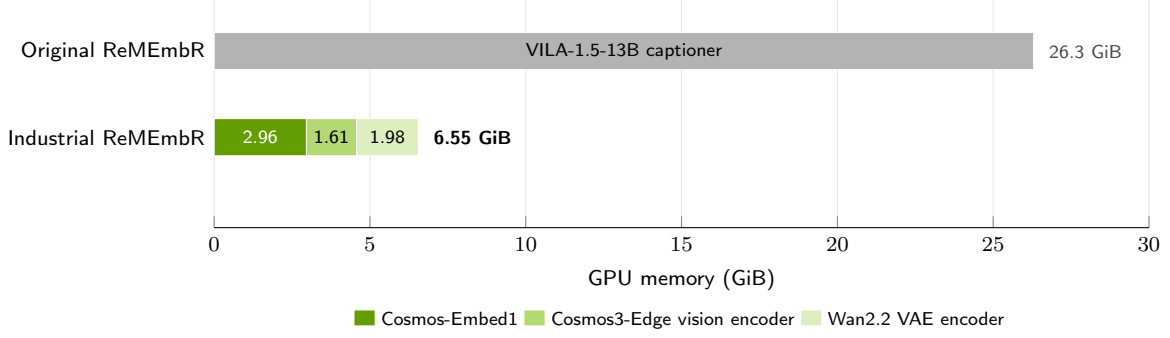
\begin{figure}[H]
    \centering
    \begin{tikzpicture}
        \begin{axis}[
            xbar stacked,
            width=0.84\linewidth,
            height=4.6cm,
            bar width=14pt,
            xmin=0, xmax=30,
            xtick={0,5,10,15,20,25,30},
            ymin=-1.05, ymax=1.6,
            xlabel={\sffamily\footnotesize GPU memory (GiB)},
            ytick={0,1},
            yticklabels={\industrialRemembr, Original ReMEmbR},
            y tick label style={font=\sffamily\footnotesize},
            xticklabel={\sffamily\footnotesize\pgfmathprintnumber{\tick}},
            axis x line*=bottom,
            axis y line*=left,
            y axis line style={draw=none},
            ytick style={draw=none},
            xmajorgrids=true,
            grid style={draw=black!10},
            set layers=standard,
            legend style={at={(0.5,-0.32)}, anchor=north, draw=none, font=\sffamily\scriptsize},
            legend columns=3,
            clip=false,
        ]
            \addplot[fill=nvidiagreen!85!black, draw=white, line width=0.4pt] coordinates {(2.96,0) (0,1)};
            \addlegendentry{Cosmos-Embed1}
            \addplot[fill=nvidiagreen!55, draw=white, line width=0.4pt] coordinates {(1.61,0) (0,1)};
            \addlegendentry{Cosmos3-Edge vision encoder}
            \addplot[fill=nvidiagreen!25, draw=white, line width=0.4pt] coordinates {(1.98,0) (0,1)};
            \addlegendentry{Wan2.2 VAE encoder}
            \addplot[fill=black!30, draw=white, line width=0.4pt, forget plot] coordinates {(0,0) (26.3,1)};

            \pgfplotsextra{
            \begin{pgfonlayer}{axis foreground}
            \node[font=\sffamily\scriptsize, text=white] at (axis cs:1.48,0) {2.96};
            \node[font=\sffamily\scriptsize] at (axis cs:3.765,0) {1.61};
            \node[font=\sffamily\scriptsize] at (axis cs:5.56,0) {1.98};
            \node[font=\sffamily\scriptsize] at (axis cs:13.15,1) {VILA-1.5-13B captioner};
            \node[font=\sffamily\scriptsize\bfseries, anchor=west] at (axis cs:6.75,0) {6.55 GiB};
            \node[font=\sffamily\scriptsize, text=black!70, anchor=west] at (axis cs:26.5,1) {26.3 GiB};
            \end{pgfonlayer}
            }
        \end{axis}
    \end{tikzpicture}
    \caption{The three iw.encode services fit in 6.55~GiB of resident GPU memory, against a 26.3~GiB sampled process peak for the original server-side captioner. The figure shows GPU-memory profiles for both (\TABREF{tab:encoder-memory}); Cosmos3-Edge loads only its vision encoder.}
    \label{fig:efficiency}
\end{figure}

\textbf{On the robot.} Robots run only encoders, and the language decoder stays on the server. The Cosmos3-Edge vision encoder takes 196~ms per frame, the Wan2.2 VAE 6.46~s for five frames in FP32, and Cosmos-Embed1 11.76~s for six frames (\TABREF{tab:xavier-latency}). Cosmos-Embed1 dominates encoding time and is the first target for engine optimization (Section~\ref{sec:road-ahead}). GPU memory is less of a constraint. The three encoder services together hold 6.55~GiB, about a quarter of the 26.3~GiB peak of the original VILA captioner (\FIGREF{fig:efficiency}).

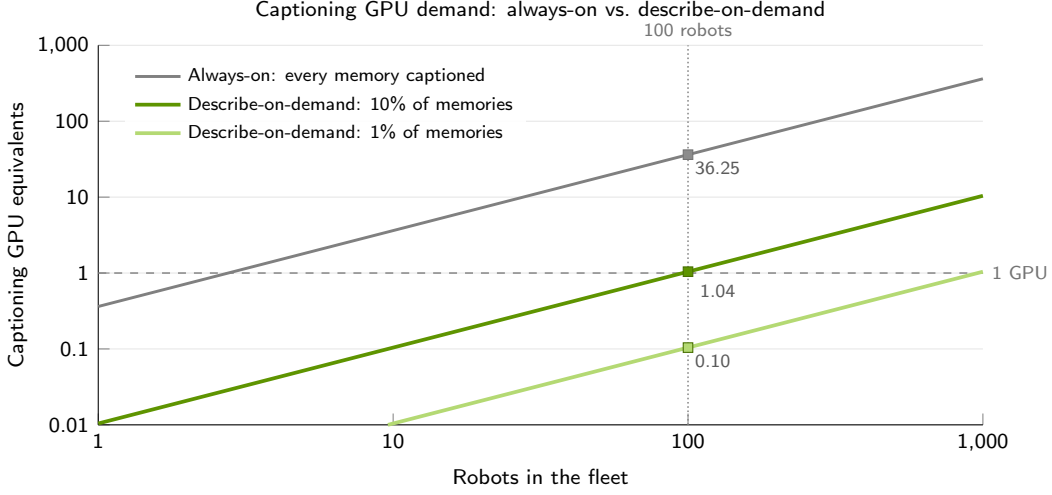
\begin{figure}[H]
    \centering
    \begin{tikzpicture}
        \begin{axis}[
            width=0.80\linewidth,
            height=6.6cm,
            title={Captioning GPU demand: always-on vs.\ describe-on-demand},
            title style={font=\sffamily\footnotesize, align=center},
            xmode=log, ymode=log,
            xmin=1, xmax=1000,
            ymin=0.01, ymax=1000,
            xlabel={\sffamily\footnotesize Robots in the fleet},
            ylabel={\sffamily\footnotesize Captioning GPU equivalents},
            ylabel style={yshift=-3pt},
            xtick={1,10,100,1000},
            xticklabels={1,10,100,{1,000}},
            ytick={0.01,0.1,1,10,100,1000},
            yticklabels={0.01,0.1,1,10,100,{1,000}},
            x tick label style={font=\sffamily\footnotesize},
            y tick label style={font=\sffamily\footnotesize},
            minor tick style={draw=none},
            axis x line*=bottom,
            axis y line*=left,
            ymajorgrids=true,
            grid style={draw=black!10},
            legend style={
                font=\sffamily\scriptsize,
                draw=none,
                fill=white,
                fill opacity=0.92,
                text opacity=1,
                at={(0.03,0.97)},
                anchor=north west,
                cells={anchor=west},
            },
            legend cell align={left},
            clip=false,
        ]
            \draw[black!45, dashed, line width=0.6pt] (axis cs:1,1) -- (axis cs:1000,1);
            \node[font=\sffamily\scriptsize, text=black!55, anchor=west] at (axis cs:1000,1) {1 GPU};
            \draw[black!45, densely dotted, line width=0.6pt] (axis cs:100,0.01) -- (axis cs:100,1000);
            \node[font=\sffamily\scriptsize, text=black!55, anchor=south] at (axis cs:100,1000) {100 robots};

            \addplot[draw=black!50, line width=1.1pt] coordinates {(1,0.3625) (1000,362.5)};
            \addlegendentry{Always-on: every memory captioned}
            \addplot[draw=nvidiagreen!80!black, line width=1.4pt] coordinates {(1,0.0104166667) (1000,10.4166667)};
            \addlegendentry{Describe-on-demand: 10\% of memories}
            \addplot[draw=nvidiagreen!55, line width=1.4pt] coordinates {(9.6,0.01) (1000,1.04166667)};
            \addlegendentry{Describe-on-demand: 1\% of memories}

            \addplot[only marks, mark=square*, mark size=1.8pt, draw=black!55, fill=black!45, forget plot] coordinates {(100,36.25)};
            \addplot[only marks, mark=square*, mark size=1.8pt, draw=nvidiagreen!60!black, fill=nvidiagreen!80!black, forget plot] coordinates {(100,1.04166667)};
            \addplot[only marks, mark=square*, mark size=1.8pt, draw=nvidiagreen!60!black, fill=nvidiagreen!55, forget plot] coordinates {(100,0.104166667)};
            \node[font=\sffamily\scriptsize, text=black!65, anchor=north west, inner sep=2.5pt] at (axis cs:100,36.25) {36.25};
            \node[font=\sffamily\scriptsize, text=black!65, anchor=west, inner sep=2.5pt] at (axis cs:104,0.58) {1.04};
            \node[font=\sffamily\scriptsize, text=black!65, anchor=north west, inner sep=2.5pt] at (axis cs:100,0.104166667) {0.10};
        \end{axis}
    \end{tikzpicture}
    \caption{Measured GPU time for captioning, extrapolated at one memory per robot every 12~s. Always-on captions every memory as written (4.35~s). Describe-on-demand captions a memory only when asked (1.25~s), for 10\% and 1\% of memories. Squares mark 100 robots.}
    \label{fig:scaling}
\end{figure}

\textbf{On the server.} Original ReMEmbR's write-time captioning takes a mean of 4.35~s per memory, whereas a description in \industrialRemembr{} costs 1.25~s from cached tokens and is generated only when requested. We measure server load in GPU equivalents, the GPU time requested per second, so one equivalent keeps one GPU fully busy. At the modeled write interval of 12~s, captioning every memory represents 36.25 GPU equivalents for 100 robots. If 10\% of memories are each described once, the serialized demand is 1.04 GPU equivalents; at 1\%, it is 0.10. At a fixed write interval and description fraction, both loads scale linearly with fleet size. Describing 10\% of memories once yields a 34.8-fold reduction. This combines the move from VILA captioning to Cosmos3-Edge decoding with deferral. Deferral costs less than captioning unless memories are described more than 3.48 times each on average. These are serialized single-GPU service-demand estimates before provisioning headroom, modeled from measured per-memory costs (\FIGREF{fig:scaling}). Appendix~\ref{app:fleet} adds the live map's agent load, sized below, and scene inspection as a symbolic term.

The agent must also keep up as the map polls more robots at once. Original ReMEmbR serves its agent through Ollama, whose throughput in our deployment barely grows with concurrency. At an 8k context, Ollama with Q8\_0 grows from 203 to 235 output tokens/s as concurrent requests rise from one to eight, while NIM in FP8 grows from 236 to 1,217~tokens/s, about five times Ollama's throughput at eight requests (\FIGREF{fig:serving-comparison}). Ollama's scheduler serves the Nemotron architecture families, including Nemotron-3-Nano-4B, one request at a time \citep{ollama-nemotron-parallelism}. We measure how many robots the agent can serve under load. With one GPU serving the agent, 95\% of map updates finish within 20~s for 24 robots polling every 10~s, and two GPUs serve at least 48 (Appendix~\ref{app:server}). The map therefore costs about 0.042 GPU equivalents per robot, or 4.2 for 100 robots. With 10\% of memories described, the server's generation load for 100 robots is about 5.2 GPU equivalents, still below the 36.25 of captioning alone, although captioning is costed unbatched here and batching it would narrow the gap. Because most of this load comes from map polling, the poll interval outweighs the description fraction as the main lever on server cost at fleet scale.

Each memory adds 0.55~MB of objects, almost all of it the 247~kB token cache and the 300~kB latent, plus a 3.2~kB index row (Appendix~\ref{app:memory}). At the modeled 12~s write interval, each robot uploads 1.3~GB per eight-hour day, an average of 0.37~Mbit/s; a fleet of 100 robots adds 131~GB of objects and 0.768~GB of index per day, or 3.9~TB and 23~GB per month (\TABREF{tab:retention}). Search reads the index and runs on the GPU. Within Milvus~\citep{wang2021milvus}, the cuVS GPU exact index~\citep{rapids2024cuvs} serves 910 exact queries per second at 32 concurrent queries, compared with 47 for CPU exact search, with identical results (\FIGREF{fig:vector-search}). Its advantage over CPU exact search grows from parity at 10,000 rows to 19-fold at one million rows; exact-search work still grows with retained rows (\FIGREF{fig:vector-search}).

\subsection{Can the Memory Keep a Live Traffic Map?}
\label{sec:results-traffic}

The live map of Section~\ref{sec:traffic} reads a traffic verdict directly from each memory's Cosmos-Embed1 vector, without a detector or a captioner. The map is only useful if this verdict is accurate and if the agent passes it on unchanged, so we compare it with caption-based memories under the same agent. The test set is a deterministic, contiguous warehouse simulation of 914 frames, each stored as a single-frame memory with its pose and time and paired with one yes/no question about traffic ahead. Labels come from the simulator's record of other agents in the zone ahead of the robot, giving 607 traffic frames and 307 clear ones. Traffic is the majority class, so always answering ``traffic'' already scores 66.4\% accuracy. We therefore report \emph{balanced accuracy}, the mean of traffic recall and clear recall, on which that baseline scores 50.0\%.

\begin{figure}[!b]
    \centering
    \begin{tikzpicture}
        \begin{axis}[
            width=0.72\linewidth,
            height=4.2cm,
            xmin=40, xmax=80,
            ymin=-0.6, ymax=3.6,
            xlabel={\sffamily\footnotesize Balanced accuracy (\%)},
            ytick={0,1,2,3},
            yticklabels={Always ``traffic'', Description + reasoning, Detection + reasoning, \textbf{Embedding verdict}},
            y tick label style={font=\sffamily\footnotesize},
            xticklabel={\sffamily\footnotesize\pgfmathprintnumber{\tick}},
            axis x line*=bottom,
            axis y line*=left,
            y axis line style={draw=none},
            ytick style={draw=none},
            ymajorgrids=true,
            grid style={draw=black!10},
            clip=false,
        ]
            \draw[black!45, dashed, line width=0.6pt] (axis cs:50,-0.6) -- (axis cs:50,3.6);
            \node[font=\sffamily\scriptsize, text=black!55, anchor=south] at (axis cs:50,3.6) {chance};

            \addplot[only marks, mark=*, mark size=2.6pt, draw=black!55, fill=black!35,
                     nodes near coords, point meta=explicit symbolic,
                     nodes near coords style={font=\sffamily\scriptsize, text=black!70, anchor=west, xshift=3pt}]
                coordinates {
                    (59.1,1) [59.1]
                    (65.3,2) [65.3]
                };
            \addplot[only marks, mark=*, mark size=2.6pt, draw=black!55, fill=black!35] coordinates {(50.0,0)};
            \addplot[only marks, mark=*, mark size=3.2pt, draw=nvidiagreen!60!black, fill=nvidiagreen!75!black,
                     nodes near coords, point meta=explicit symbolic,
                     nodes near coords style={font=\sffamily\scriptsize, text=black!80, anchor=west, xshift=3pt}]
                coordinates {(71.3,3) [\textbf{71.3}]};

            \node[font=\sffamily\scriptsize, text=black!55, anchor=south] at (axis cs:86.5,3.6) {frames scored};
            \foreach \y/\n in {3/914, 2/844, 1/902, 0/914} {
                \edef\temp{\noexpand\node[font=\noexpand\sffamily\noexpand\scriptsize, text=black!70] at (axis cs:86.5,\y) {\n\noexpand\,/\noexpand\,914};}
                \temp
            }
        \end{axis}
    \end{tikzpicture}
    \caption{The caption-free embedding verdict (green) has the highest balanced accuracy among the reported representations. Results are on simulated warehouse traffic over 914 frames, by what each frame writes to memory; caption arms use reasoning. Scores exclude frames without usable context; coverage is listed at the right. The dashed line marks chance, where always answering ``traffic'' sits.}
    \label{fig:traffic-results}
\end{figure}
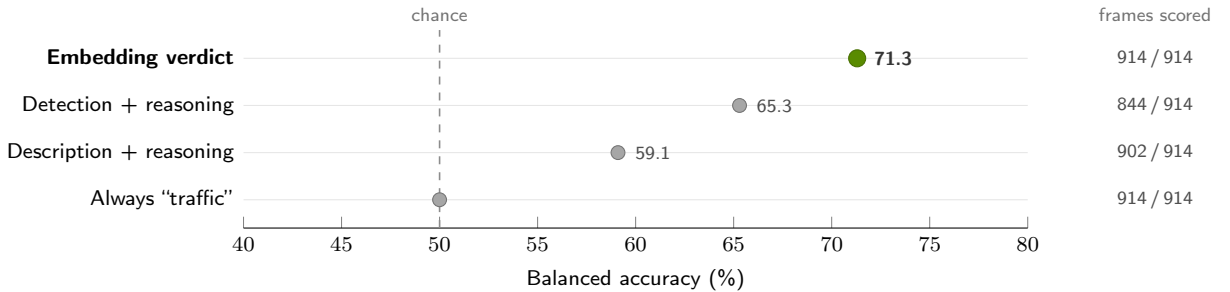

The three memories differ only in what each frame writes. The \emph{embedding verdict} scores the frame's Cosmos-Embed1 vector against the occupied and clear prototypes of Section~\ref{sec:reporting}, with no tuned threshold. For \emph{detection}, Cosmos3-Edge lists the obstacles in the forward path as labeled bounding boxes, and for \emph{description} it writes a free-text description of the scene, both with reasoning enabled. Before answering, the agent must search memory by the frame's position, so it reads every representation from the same memory row. A frame counts toward a representation's \emph{coverage} when that representation produced usable content for it. Prompts and prototypes are listed in Appendix~\ref{app:prompts}, and protocols in Appendix~\ref{app:traffic}.

The embedding verdict reaches 71.3\% balanced accuracy, 6.0 points above detection, the best caption-based memory, and it is the only representation that covers all 914 frames, against 844 for detection and 902 for description (\FIGREF{fig:traffic-results}). The caption arms are scored only on the frames they cover, so counting dropped frames as errors would only widen this margin. The two caption-based memories fail in opposite directions. Detection finds traffic in 88.4\% of traffic frames but recognizes only 42.3\% of clear frames, so its 72.7\% accuracy, slightly above the verdict's 72.1\%, mostly reflects the majority class. Description errs the other way, with 91.1\% clear recall and only 27.1\% traffic recall. The verdict balances the two classes, with 73.8\% traffic recall and 68.7\% clear recall (\TABREF{tab:traffic}).

Routing the verdict through the agent barely changes its classification. Nemotron agrees with the raw embedding verdict on 913 of 914 frames, scoring 659/914 against the direct classifier's 660/914. It retrieves the target memory in 99.6\% of cases and produces no unparseable answers. The map can keep the agent as its interface, which lets it ask new questions of the same memory.

\section{Road Ahead}
\label{sec:road-ahead}

The traffic map is the first application built on fleet memory. The directions below reuse the same encoding and storage, and several of them require only new tools on the server.

\textbf{New uses of the same memory.} The agent can already locate any landmark that any robot has seen, yet it uses that knowledge only to answer questions. Recast as semantic subgoals, such as ``pass the blue loading bay, then turn left at the marked crossing,'' that knowledge could guide a robot along routes it has never driven, drawing on what other robots have already observed. The live map's zone verdicts can play the same role for fleet scheduling. RobotFleet~\citep{gupta2025robotfleet} allocates subtasks across a heterogeneous fleet from a shared world state of declarative statements and replans when a robot reports something new. An occupied zone can update that state as soon as its memory arrives, and the agent can be asked what is blocking the zone before the next allocation.

\textbf{Fleet memory as training data.} Each robot already encodes its clips into the vision tokens and video latents that compatible Cosmos3 models consume~\citep{nvidia2026cosmos3}, so routine fleet operation doubles as distributed preprocessing for training. This costly visual encoding is offloaded to edge devices across the fleet, paid once per memory, and amortized over every epoch and every compatible model trained from the cached representations, without transferring raw video or re-running the encoders (Appendix~\ref{app:finetune}). A world model trained on these latents could become a learned simulator of the site, replaying recorded scenes under alternative robot behaviors to show what would have happened had a robot taken another aisle, therefore translating failed rollouts into successful ones and distilling back to deployed policies and closing the flywheel. Fine-tuning the language decoder on the cached vision tokens could teach it the site's own objects and terminology. Fine-tuning Cosmos-Embed1 with location-aware captioning objectives, as in LocCa~\citep{wan2024locca} and SigLIP~2~\citep{tschannen2025siglip2}, could also strengthen the spatial awareness that physical AI tasks require, thereby improving the balanced accuracy on traffic verdicts.

\textbf{A cheaper memory at every stage.} On the robot, the general-purpose video--text embedder answers the traffic question with two dot products but dominates encoding time. An embedder specialized for industrial scenes and distilled into a compact backbone such as SigLIP~2~\citep{tschannen2025siglip2} could encode faster on the edge and sharpen every prompt-based check on factory and warehouse vocabulary. On the server, the edge deployment expertise behind our encoders could yield a leaner agent, pruned, quantized and distilled from open models, that answers map polls with fewer calls and tokens. In storage, every memory is kept in full, so the store grows linearly, reaching 3.9~TB per month for 100 robots at the modeled write interval (\TABREF{tab:retention}). Much like human episodic memory, fleet memory could consolidate routine observations of the same place while keeping surprises in full, so that storage grows with the site's rate of change rather than with fleet operating hours.

\textbf{The physical world as an attack surface.} Text that the decoder reads from a scene enters the agent's context like any other content, so a sign or a screen in view of a camera becomes an input the agent might obey, a physical form of prompt injection. As the agent gains authority over the fleet, perception must be treated as untrusted data. iw.shield, a protection layer under development, is our first step toward agents that can read the world without taking malicious orders from it.
\section{Conclusion}
\label{sec:conclusion}
We present \industrialRemembr{}, a shared fleet memory that decouples observation encoding from on-demand interpretation. Each robot encodes a clip once into a retrieval embedding, vision tokens, and a video latent while retaining raw camera frames onboard. These representations support semantic retrieval, live traffic mapping, scene description, and reconstruction. We derive traffic classifications directly from retrieval embeddings at encoding time and generate scene descriptions and reconstructions only when needed.
Encoding, storage, mapping, and interpretation operate as independent services, allowing multiple applications to reuse each observation and individual models to be replaced independently.

Our evaluation demonstrates the benefits of deferred description and validates key components of the architecture. Deferring caption generation removes text generation from the memory-write path.
Video-based indexing matches or improves retrieval recall in every matched comparison, and cached-token decoding reproduces the unsplit Cosmos3-Edge output in all 450 tests. On full NaVQA, the updated model stack reduces first-attempt parse failures from 20.7\% to 13.8\% and unparsable answers from 4.5\% to 1.9\% with an agent half the parameter size. At a modeled 12~s write interval, description demand for 100 robots is 1.04 GPU equivalents when 10\% of memories are described, compared with 36.25 GPU equivalents for captioning every memory. Across 914 simulated warehouse frames, embedding-based traffic classification achieves 71.3\% balanced accuracy, exceeding the best reasoning-enabled caption-based alternative by 6.0 percentage points.

Further engineering is required for production deployment. Cosmos-Embed1, the slowest encoder on Jetson AGX Xavier, is our first target for TensorRT optimization and fine-tuning, while agent provisioning must account for the map-update workload and polling cadence. We plan to extend the system to congestion-aware fleet planning, semantic guidance for navigation policies, and continuous curation of deployment data for model fine-tuning. We will also investigate deploying agents on robot edge devices, reorganizing encoding components on the robot to obfuscate people from memories directly on the robots' volatile memory, and developing iw.shield to protect the system against cyberattacks. The broader contribution is a reusable memory architecture through which a fleet's collective experience can support an expanding range of applications as models, tasks, and warehouse conditions evolve.

\newpage
\bibliographystyle{plainnat}
\bibliography{main}

\newpage
\appendix
\section*{Appendix}

Appendix~\ref{app:impl} describes the memory record and the agent. Appendices~\ref{app:navqa}--\ref{app:traffic} give the protocols and full results behind Sections~\ref{sec:results-navqa}--\ref{sec:results-traffic}, Appendix~\ref{app:prompts} lists the prompts, and Appendix~\ref{app:finetune} outlines training on the fleet cache (Section~\ref{sec:road-ahead}).

\section{System Reference}
\label{app:impl}

\subsection{Memory Record}
\label{app:memory}

A memory covers a three-second clip sampled at 2~FPS. NaVQA uses contiguous clips, and the fleet model of Appendix~\ref{app:fleet} assumes one clip every 12~s. \TABREF{tab:encoders} lists the stored representations. The Wan2.2 VAE runs in FP32, and its latent is cast to BF16 for storage, halving it from 599~kB to 300~kB.

\begin{table}[H]
    \centering
    \footnotesize
    \setlength{\tabcolsep}{6pt}
    \begin{tabular}{@{}llllr@{}}
        \toprule
        Use & Encoder & Input ($T \times H \times W$) & Stored tensor & Size (kB) \\
        \midrule
        Retrieval   & Cosmos-Embed1       & $6 \times 448 \times 448$ & $768$, FP32                             & 3.1 \\
        Description & Cosmos3-Edge vision & $1 \times 192 \times 320$ & $60 \times 2048$, BF16                  & 246.6 \\
        Inspection  & Wan2.2 VAE          & $5 \times 480 \times 832$ & $48 \times 2 \times 30 \times 52$, BF16 & 299.5 \\
        \bottomrule
    \end{tabular}
    \caption{Memory record for one three-second clip. Each memory retains 546.1~kB of objects and uploads 549.2~kB with the search vector; its index row is 3.2~kB. The vision-token size includes the token-grid and prompt-template metadata the decoder reads.}
    \label{tab:encoders}
\end{table}

\textbf{Storage and index.} Each memory is one Milvus row holding the search vector, the position $(x, y, z)$, the heading, the capture time and the traffic verdict. Vision tokens and latents sit in an S3-compatible object store (SeaweedFS) under a per-memory prefix that the row references. All search is exact: vectors use the embedder's similarity metric, and position and time use Euclidean distance. The accuracy runs use CPU search, and Appendix~\ref{app:server} compares it with the cuVS~\citep{rapids2024cuvs} \texttt{GPU\_BRUTE\_FORCE} index in Milvus~\citep{wang2021milvus}.

\subsection{Agent, Tools and Services}
\label{app:agent}

\textbf{Agent loop.} Nemotron-3-Nano-4B may call tools for up to three rounds, with several calls per round, before a final step writes a structured JSON answer with text, yes/no, position, time and duration fields. Each search returns the $k$ nearest memories ($k = 5$ in the NaVQA runs). An answer that fails to parse is regenerated, up to three attempts in total. In deployment, NVIDIA NIM serves the agent in FP8 with a 16k-token context and a 4,096-token output limit, and removes the reasoning trace before tool calls are parsed. The describe tool decodes greedily from the cached tokens with at most 512 new tokens. Tool signatures and the answer fields are listed in Appendix~\ref{app:prompts-agent}.

\textbf{Services.} Each model service holds one loaded instance and returns HTTP~503 when the model stays busy for more than one second or runs out of GPU memory, rather than queueing without bound.

\textbf{Live semantic map.} Zones are directed floor-plan polygons at 0.05~m per pixel. Every 10~s, the map asks the agent, once for each robot it has seen, whether there is traffic at that robot's position and colors the zone as traffic or clear. The agent runs with reasoning enabled.

\section{NaVQA Protocol and Full Results}
\label{app:navqa}

\textbf{Protocol.} The ablation covers all seven CODa sequences (0, 3, 4, 6, 16, 21 and 22) with three trials per configuration; the yes/no column counts 67 binary questions per trial, 201 responses per configuration. The indicated captioner writes the descriptions; Cosmos-Embed1 indexes the video, whereas MxBai indexes the caption text. Agents run with reasoning disabled, and malformed yes/no responses count as incorrect.

\textbf{All configurations.} \TABREF{tab:grid} reports every combination of agent, captioner and index. Across matched swaps, which change one component and hold the other two fixed, the video index never lowers recall (+0.2 to +8.5 points). The Nemotron agent lowers first-attempt failures in all four settings (by 4.5 to 10.5 points) and unparsable answers by 2.4 to 4.4 points, at a cost of 2.3 to 5.9 points of recall. The captioner swap has no consistent direction.

\begin{table}[H]
    \centering
    \footnotesize
    \setlength{\tabcolsep}{3pt}
    \begin{tabular}{@{}lllccccccl@{}}
        \toprule
        Agent & Captioner & Index & First-try & Unparsable & Recall & Position & Time & Yes/no & System \\
              &           &       & fail.\ (\%) $\downarrow$ & (\%) $\downarrow$ & (\%) $\uparrow$ & err.\ (m) $\downarrow$ & err.\ (min) $\downarrow$ & ok $\uparrow$ & \\
        \midrule
        Llama    & VILA         & MxBai         & 20.7 & 4.5 & 22.9 & 69.1 & 1.64 & 87/201  & Original \\
        Llama    & VILA         & Cosmos-Embed1 & 23.5 & 5.5 & 29.7 & 79.4 & 2.17 & 94/201  & \\
        Llama    & Cosmos3-Edge & MxBai         & 22.4 & 3.6 & 26.7 & 74.6 & 3.38 & 109/201 & \\
        Llama    & Cosmos3-Edge & Cosmos-Embed1 & 22.0 & 4.3 & 26.9 & 76.2 & 2.19 & 106/201 & \\
        \midrule
        Nemotron & VILA         & MxBai         & 16.2 & 0.9 & 18.7 & 71.0 & 2.21 & 81/201  & \\
        Nemotron & VILA         & Cosmos-Embed1 & 13.0 & 1.1 & 27.2 & 67.5 & 1.87 & 74/201  & \\
        Nemotron & Cosmos3-Edge & MxBai         & 12.2 & 0.9 & 20.8 & 66.9 & 1.66 & 81/201  & \\
        Nemotron & Cosmos3-Edge & Cosmos-Embed1 & 13.8 & 1.9 & 24.6 & 66.6 & 2.37 & 81/201  & Upgraded \\
        \bottomrule
    \end{tabular}
    \caption{Full seven-sequence NaVQA component ablation: three trials, 201 yes/no responses per configuration. Agents are Llama-3.1-8B and Nemotron-3-Nano-4B; VILA is VILA-1.5-13B. ``First-try fail.'' is the first-attempt parse-failure rate. Position and time errors are medians over parsed answers. ``Original'' is original ReMEmbR and ``Upgraded'' is \stackRemembr.}
    \label{tab:grid}
\end{table}

\section{Describe-on-Demand and Reconstruction}
\label{app:describe}

\textbf{Describe-on-demand run.} In a separate 150-frame run, each frame is encoded once by the Cosmos3-Edge vision encoder, and its cached tokens are decoded with three prompts: the detection prompt of Appendix~\ref{app:prompts-traffic} (at most 256 new tokens), a general scene-description request (512) and a one-word yes/no question (16). Decoding is greedy with reasoning disabled, and each prompt gets a single attempt. The same frame and prompt are also run through the unsplit model. Two outputs match when the decoded strings are equal after trimming surrounding whitespace; all 450 frame--prompt pairs match. Times are wall-clock after one warm-up and include tokenization and image preprocessing.

\textbf{Reconstruction run.} Each of the 150 traffic frames is repeated five times into a $5 \times 480 \times 832$ input, encoded, and reconstructed from its stored BF16 latent; reconstruction time is a median over frames.

\textbf{Latent precision.} Decoding the BF16 latent in place of the FP32 one changes the reconstruction metrics of Section~\ref{sec:results-describing} by less than 0.1\% of their value (\TABREF{tab:bf16}).

\begin{table}[H]
    \centering
    \footnotesize
    \setlength{\tabcolsep}{10pt}
    \begin{tabular}{@{}lrrr@{}}
        \toprule
        Metric & BF16 latent & Change vs.\ FP32 & Relative change \\
        \midrule
        PSNR (dB) $\uparrow$  & 39.7   & $-0.003$             & 0.008\% \\
        SSIM $\uparrow$       & 0.974  & $-1.3 \times 10^{-5}$ & 0.001\% \\
        LPIPS $\downarrow$    & 0.0052 & $+4 \times 10^{-6}$   & 0.08\% \\
        \bottomrule
    \end{tabular}
    \caption{Reconstruction fidelity of the stored BF16 latent over 150 traffic frames, each repeated five times, and its change from decoding the FP32 latent.}
    \label{tab:bf16}
\end{table}

\section{Robot, Server and Fleet Costs}
\label{app:costs}

\subsection{Jetson AGX Xavier Engine Profiling}
\label{app:xavier}

\TABREF{tab:xavier-latency} gives the encoder latencies behind Section~\ref{sec:results-cost}.

\begin{table}[H]
\centering\footnotesize
\begin{tabular}{@{}llrrr@{}}
\toprule
Encoder & Input ($T \times H \times W$) & Mean latency (s) & P95 latency (s) & Runs \\
\midrule
Cosmos-Embed1 & $6 \times 448 \times 448$ & 11.76 & 11.78 & 10 \\
Cosmos3-Edge vision & $1 \times 192 \times 320$ & 0.196 & 0.254 & 16 \\
Wan2.2 VAE (FP32) & $5 \times 480 \times 832$ & 6.46 & 6.51 & 10 \\
\bottomrule
\end{tabular}
\caption{Mean and P95 latency of warmed TensorRT~8.5.2 engines on a Jetson AGX Xavier with random inputs, including host--device transfers. Cosmos-Embed1 uses custom TensorRT plugins.}
\label{tab:xavier-latency}
\end{table}

\subsection{Server Memory, Serving and Search}
\label{app:server}

\textbf{GPU memory.} \TABREF{tab:encoder-memory} lists the GPU memory of each process in \FIGREF{fig:efficiency}.

\begin{table}[H]
    \centering
    \footnotesize
    \setlength{\tabcolsep}{8pt}
    \begin{tabular}{@{}lr@{}}
        \toprule
        Process & GPU memory (GiB) \\
        \midrule
        \multicolumn{2}{@{}l}{\emph{Robot (iw.encode)}} \\
        Cosmos-Embed1                 & 2.96 \\
        Cosmos3-Edge vision encoder   & 1.61 \\
        Wan2.2 VAE encoder            & 1.98 \\
        \textbf{All three loaded}     & \textbf{6.55} \\
        \midrule
        \multicolumn{2}{@{}l}{\emph{Server}} \\
        Cosmos3-Edge language decoder & 5.4 \\
        Wan2.2 VAE decoder            & 10.3 \\
        Nemotron-3-Nano-4B (NIM, FP8) & 5.9 \\
        \midrule
        \multicolumn{2}{@{}l}{\emph{Original ReMEmbR}} \\
        VILA-1.5-13B captioner        & 26.3 \\
        \bottomrule
    \end{tabular}
    \caption{GPU memory by process. Encoder values are resident memory with the services loaded; server and captioner values are peaks during inference; the NIM value excludes the key-value cache that NIM reserves.}
    \label{tab:encoder-memory}
\end{table}

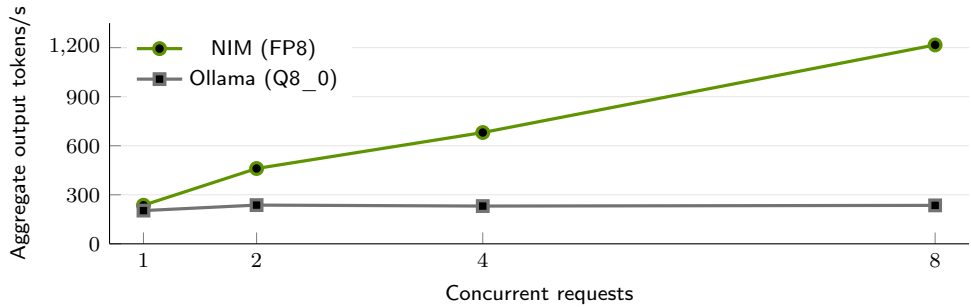
\begin{figure}[H]
    \centering
    \begin{tikzpicture}
        \begin{axis}[
            width=0.78\linewidth,
            height=4.5cm,
            xmin=0.7, xmax=8.3,
            ymin=0, ymax=1350,
            xlabel={\sffamily\footnotesize Concurrent requests},
            ylabel={\sffamily\footnotesize Aggregate output tokens/s},
            xtick={1,2,4,8},
            ytick={0,300,600,900,1200},
            x tick label style={font=\sffamily\footnotesize},
            y tick label style={font=\sffamily\footnotesize},
            minor tick style={draw=none},
            axis x line*=bottom,
            axis y line*=left,
            ymajorgrids=true,
            grid style={draw=black!10},
            legend style={font=\sffamily\footnotesize, draw=none, at={(0.02,0.98)}, anchor=north west},
            clip=false,
        ]
            \addplot[draw=nvidiagreen!80!black, mark=*, mark size=2.2pt, line width=1.2pt]
                coordinates {(1,236.1) (2,460.7) (4,680.7) (8,1216.6)};
            \addlegendentry{NIM (FP8)}
            \addplot[draw=black!55, mark=square*, mark size=2.0pt, line width=1.2pt]
                coordinates {(1,203.2) (2,236.9) (4,231.0) (8,235.2)};
            \addlegendentry{Ollama (Q8\_0)}
        \end{axis}
    \end{tikzpicture}
    \caption{NIM (FP8) scales to 1217 tokens/s at 8 concurrent requests, while Ollama (Q8\_0) stays near 235. Throughput is aggregate output for Nemotron-3-Nano-4B at 8k context on one RTX PRO 6000 Blackwell Server Edition.}
    \label{fig:serving-comparison}
\end{figure}

\textbf{Agent serving.} \FIGREF{fig:serving-comparison} compares NIM (FP8) and Ollama (Q8\_0) serving Nemotron-3-Nano-4B on synthetic 8k-context requests at concurrency 1 to 8. A separate sweep (\TABREF{tab:nim}) uses a larger NIM memory allocation and extends concurrency to 256. Throughput peaks at 2,844 output tokens/s at 128 concurrent requests; beyond that, time to first token rises sharply. All requests completed without errors.

\begin{table}[H]
    \centering
    \footnotesize
    \setlength{\tabcolsep}{6pt}
    \begin{tabular}{@{}rrrrr@{}}
        \toprule
        Concurrency & Output tok/s & Mean TPOT (ms) & Mean TTFT (s) & P99 TTFT (s) \\
        \midrule
        1   & 155.8  & 5.96  & 0.097  & 0.120 \\
        2   & 297.3  & 6.11  & 0.149  & 0.167 \\
        4   & 570.2  & 6.42  & 0.272  & 0.307 \\
        8   & 953.3  & 7.21  & 0.447  & 0.541 \\
        16  & 1574.7 & 8.71  & 0.715  & 1.032 \\
        32  & 2026.9 & 12.52 & 1.002  & 1.674 \\
        48  & 2357.4 & 16.59 & 1.131  & 1.735 \\
        64  & 2670.4 & 20.62 & 1.079  & 1.412 \\
        96  & 2726.6 & 29.19 & 1.085  & 1.314 \\
        128 & 2844.1 & 37.78 & 1.139  & 1.220 \\
        192 & 2252.1 & 57.59 & 1.115  & 4.660 \\
        256 & 1909.9 & 64.91 & 14.173 & 28.339 \\
        \bottomrule
    \end{tabular}
    \caption{NIM FP8 synthetic-request sweep. TTFT is time to first token; TPOT is time per output token.}
    \label{tab:nim}
\end{table}

\textbf{Live-map capacity.} In the load test, each emulated robot requests one map update every 10~s. Each update makes two to three agent calls, with input and output lengths similar to those of the live map. A setup serves a fleet when 95\% of its updates finish within 20~s (\TABREF{tab:map-capacity}). In the live map itself, 17 robots on one GPU see a mean update time of 4.5~s.

\begin{table}[H]
    \centering
    \footnotesize
    \setlength{\tabcolsep}{10pt}
    \begin{tabular}{@{}lrr@{}}
        \toprule
        Robots & 1 GPU & 2 GPUs \\
        \midrule
        8  & 8.1   & 7.1  \\
        16 & 11.2  & 8.2  \\
        20 & 13.3  & 8.9  \\
        24 & 17.0  & 9.5  \\
        28 & 23.6  & 11.4 \\
        32 & 37.2  & 11.6 \\
        40 & 76.7  & 13.8 \\
        48 & 116.2 & 17.6 \\
        \midrule
        Robots served & \textbf{24} & \textbf{48} \\
        \bottomrule
    \end{tabular}
    \caption{P95 map-update time (s) for robots polling every 10~s, with one NIM instance per GPU. A setup serves a fleet when 95\% of updates finish within 20~s; 48 robots is the largest fleet tested.}
    \label{tab:map-capacity}
\end{table}

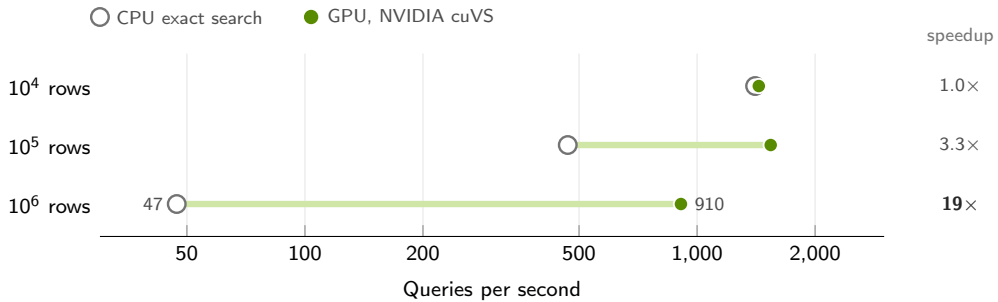
\begin{figure}[H]
    \centering
    \begin{tikzpicture}
        \begin{axis}[
            width=0.72\linewidth,
            height=4.0cm,
            xmode=log,
            xmin=30, xmax=3000,
            ymin=-0.55, ymax=2.55,
            xlabel={\sffamily\footnotesize Queries per second},
            xtick={50,100,200,500,1000,2000},
            xticklabels={50,100,200,500,{1,000},{2,000}},
            x tick label style={font=\sffamily\footnotesize},
            ytick={0,1,2},
            yticklabels={{$\mathsf{10^6}$ rows},{$\mathsf{10^5}$ rows},{$\mathsf{10^4}$ rows}},
            y tick label style={font=\sffamily\footnotesize},
            minor tick style={draw=none},
            axis x line*=bottom,
            axis y line*=left,
            y axis line style={draw=none},
            ytick style={draw=none},
            xmajorgrids=true,
            grid style={draw=black!10},
            clip=false,
        ]
            \draw[nvidiagreen!35, line width=2.4pt, line cap=round] (axis cs:1406,2) -- (axis cs:1437,2);
            \draw[nvidiagreen!35, line width=2.4pt, line cap=round] (axis cs:468,1) -- (axis cs:1541,1);
            \draw[nvidiagreen!35, line width=2.4pt, line cap=round] (axis cs:47.1,0) -- (axis cs:909.7,0);

            \addplot[only marks, mark=*, mark size=3.3pt,
                     mark options={draw=black!55, fill=white, line width=0.9pt}]
                coordinates {(1406,2) (468,1) (47.1,0)};
            \addplot[only marks, mark=*, mark size=2.7pt,
                     mark options={draw=white, fill=nvidiagreen!75!black, line width=0.7pt}]
                coordinates {(1437,2) (1541,1) (909.7,0)};

            \node[font=\sffamily\scriptsize, text=black!70, anchor=east, inner sep=5pt] at (axis cs:47.1,0) {47};
            \node[font=\sffamily\scriptsize, text=black!70, anchor=west, inner sep=5pt] at (axis cs:909.7,0) {910};

            \node[font=\sffamily\scriptsize, text=black!55, anchor=south] at (axis cs:4700,2.55) {speedup};
            \node[font=\sffamily\scriptsize, text=black!70] at (axis cs:4700,2) {1.0$\times$};
            \node[font=\sffamily\scriptsize, text=black!70] at (axis cs:4700,1) {3.3$\times$};
            \node[font=\sffamily\scriptsize\bfseries, text=black!85] at (axis cs:4700,0) {19$\times$};

            \draw[draw=black!55, fill=white, line width=0.9pt] (rel axis cs:0.0,1.2) circle (3.3pt);
            \node[font=\sffamily\scriptsize, text=black!70, anchor=west, inner sep=6pt] (lcpu) at (rel axis cs:0.0,1.2) {CPU exact search};
            \fill[nvidiagreen!75!black] ([xshift=12pt]lcpu.east) circle (2.7pt);
            \node[font=\sffamily\scriptsize, text=black!70, anchor=west, inner sep=6pt] at ([xshift=12pt]lcpu.east) {GPU, NVIDIA cuVS};
        \end{axis}
    \end{tikzpicture}
    \caption{GPU exact search's throughput advantage over CPU exact search grows across the tested index sizes, from parity at $10^4$ rows to 3.3$\times$ at $10^5$ and 19$\times$ at $10^6$. Results are for 32 concurrent queries in one Milvus instance over synthetic 768-dimensional vectors; the GPU index uses NVIDIA cuVS~\citep{rapids2024cuvs}. Both searches return identical neighbours. The axis is logarithmic, so the length of each green bar is the relative speedup.}
    \label{fig:vector-search}
\end{figure}

\textbf{Vector search.} At one million rows, a single query takes 10.6~ms with the cuVS GPU index and 22.8~ms on the CPU (\FIGREF{fig:vector-search}).

\subsection{Fleet Model}
\label{app:fleet}

\textbf{Notation.} Let $N$ be the number of robots and $r = 1/12$~s$^{-1}$ the write rate of one robot, one memory every 12~s. An operating day is $h_{\mathrm{op}} = 28{,}800$~s (8~h), so a robot writes 2{,}400 memories per day. Service times are $t_{\mathrm{cap}} = 4.35$~s for original captioning and $t_{\mathrm{desc}} = 1.25$~s for describing a memory from cached tokens, both means, and $t_{\mathrm{insp}}$ for scene inspection, kept symbolic. Per new memory, $p$ is the mean number of descriptions and $q$ the mean number of inspections; $p = 0.1$ and $p = 0.01$ mean one description for 10\% and 1\% of memories. The live map polls every $\tau$~s (10~s by default), and $u_a$ is the rate of map updates one GPU completes while answering 95\% of them within two polling periods, 20~s. One GPU serves at least 24 robots within this limit (\TABREF{tab:map-capacity}), so $u_a \approx 2.4$ updates/s. Unlike the serialized description terms, $u_a$ is a batched serving rate. A memory keeps $b = 546.1$~kB of objects and a 3.2~kB index row, and $D$ is the retention in days. A GPU equivalent is the serialized service time offered per second, so one equivalent is one fully busy GPU.

\textbf{Per-robot demand.} \TABREF{tab:unitcosts} gives the demand of one robot; all quantities scale linearly with $N$.

\begin{table}[H]
    \centering
    \footnotesize
    \setlength{\tabcolsep}{7pt}
    \begin{tabular}{@{}llll@{}}
        \toprule
        Event & Input & Rate & Per robot \\
        \midrule
        Caption a memory (original) & $t_{\mathrm{cap}} = 4.35$~s     & $r$      & $0.3625$ GPU eq. \\
        Describe a memory           & $t_{\mathrm{desc}} = 1.25$~s    & $p\,r$   & $0.104\,p$ GPU eq. \\
        Inspect a scene             & $t_{\mathrm{insp}}$             & $q\,r$   & $q\,t_{\mathrm{insp}}/12$ GPU eq. \\
        Update the map              & $u_a \approx 2.4$ updates/s/GPU & $1/\tau$ & $0.042$ GPU eq. \\
        Retain objects             & $b = 546.1$~kB               & $r$      & $1.31\,D$ GB \\
        Index rows                 & 3.2~kB                          & $r$      & $7.68\,D$ MB \\
        \bottomrule
    \end{tabular}
    \caption{Per-robot demand and storage. Sources are \TABREF{tab:ondemand}, \TABREF{tab:encoders} and \TABREF{tab:map-capacity}.}
    \label{tab:unitcosts}
\end{table}

\textbf{Cost model.} Eq.~\ref{eq:cost-model} gives the monthly server GPU and object-storage cost of one robot for a retention of $D$ days,
\begin{equation}
    C_{\mathrm{robot}} = C_g\left[\frac{1}{\tau u_a} + r\,(p\,t_{\mathrm{desc}} + q\,t_{\mathrm{insp}})\right] + C_s\,\frac{r\,h_{\mathrm{op}}\,D\,b}{10^6},
    \label{eq:cost-model}
\end{equation}
where $C_g$ is the monthly cost of one reserved GPU, $C_s$ the storage cost per GB-month and $b$ is in kB. Original write-time captioning alone costs $0.3625\,C_g$ per robot. At $\tau = 10$~s and $p = 0.1$, the map and description terms together cost $0.052\,C_g$ per robot, so even with the live map, which original ReMEmbR lacks, server generation demand stays below that of captioning alone.

\textbf{Break-even and the 34.8$\times$ factor.} Describing from cached tokens offers less generation demand than write-time captioning whenever $p\,t_{\mathrm{desc}} < t_{\mathrm{cap}}$, that is, below $p < t_{\mathrm{cap}}/t_{\mathrm{desc}} = 4.35/1.25 = 3.48$ descriptions per memory. At $p = 0.1$ and $p = 0.01$ the demand falls by $t_{\mathrm{cap}}/(p\,t_{\mathrm{desc}}) = 34.8$ and $348$ times at any fleet size. These factors combine two changes: replacing VILA captioning with Cosmos3-Edge decoding, and deferring description to the memories that are asked about.

\textbf{Fleet projection and retention.} \TABREF{tab:retention} projects upload and retention for one robot and a 100-robot fleet. One robot uploads 549~kB every 12~s, an average of 0.37~Mbit/s while writing.

\begin{table}[H]
    \centering
    \footnotesize
    \setlength{\tabcolsep}{8pt}
    \begin{tabular}{@{}llrrrr@{}}
        \toprule
        Scope & Period & Memories & Uploaded & Objects & Index \\
        \midrule
        1 robot    & 1 day    & 2,400      & 1.32 GB   & 1.31 GB  & 7.68 MB \\
        \midrule
        100 robots & 1 day    & 240,000    & 131.81 GB & 131.07 GB & 0.768 GB \\
        100 robots & 30 days  & 7.2 million & 3.95 TB  & 3.93 TB  & 23.04 GB \\
        100 robots & 365 days & 87.6 million & 48.11 TB & 47.84 TB & 280.32 GB \\
        \bottomrule
    \end{tabular}
    \caption{Totals over each period at one memory every 12~s over an 8-hour day.}
    \label{tab:retention}
\end{table}

\section{Traffic Evaluation}
\label{app:traffic}

\textbf{Dataset.} The evaluation set is 914 consecutive RGB frames of one warehouse simulation run from a single ego robot. Of the 607 traffic frames, 289 are low severity and 318 are high; 307 frames are clear.

\textbf{Embedding verdict.} Each class has an ensemble of short text prompts (Appendix~\ref{app:prompts-traffic}). Their Cosmos-Embed1 text embeddings are mean-pooled and renormalized into one vector per class, and a clip is labeled by whichever class vector its video embedding is closer to. The verdict is written to memory as one sentence.

\textbf{Caption arms.} The Cosmos3-Edge detection and description arms run at temperature~0 with reasoning enabled and 2,048 output tokens each. Detection returns one JSON array of \texttt{\{label, bbox\_2d\}} objects in normalized 0 to 1000 coordinates. A response that does not parse, or has the wrong keys, a generic label or an out-of-range or misordered box, is re-asked with the shorter correction prompt, for up to three attempts in total, and the frame is dropped if all three fail. Severity follows from the number of validated boxes: none is clear, one is low and two or more is high. Description asks for one sentence of at most 40 words. A description whose reasoning was cut off, or that exceeds the length limit, is dropped. The prompts are in Appendix~\ref{app:prompts-traffic}.

\textbf{Full results.} \TABREF{tab:traffic} gives coverage, accuracy, balanced accuracy and per-class recall, split by severity, for every representation in \FIGREF{fig:traffic-results}. Detection has the highest recall at both traffic severities, while the embedding verdict retains more balanced traffic and clear recall.

\begin{table}[H]
    \centering
    \footnotesize
    \setlength{\tabcolsep}{5pt}
    \begin{tabular}{@{}lcccccccc@{}}
        \toprule
        Memory per frame & Coverage & Accuracy & Balanced & \multicolumn{3}{c}{Traffic recall} & Clear \\
        \cmidrule(lr){5-7}
                         &          &          & acc.     & All & Low & High & recall \\
        \midrule
        Embedding verdict       & \textbf{914/914} & 72.1          & \textbf{71.3} & 73.8          & 68.5          & 78.6          & 68.7 \\
        Detection + reasoning   & 844/914          & \textbf{72.7} & 65.3          & \textbf{88.4} & \textbf{87.0} & \textbf{89.6} & 42.3 \\
        Description + reasoning & 902/914          & 48.7          & 59.1          & 27.1          & 25.1          & 28.9          & \textbf{91.1} \\
        \midrule
        Always ``traffic''      & 914/914          & 66.4          & 50.0          & 100.0         & 100.0         & 100.0         & 0.0 \\
        \bottomrule
    \end{tabular}
    \caption{Simulated warehouse traffic over 914 frames (\%). The embedding verdict is read from the Cosmos-Embed1 vector; the detection and description arms use Cosmos3-Edge with reasoning enabled. Coverage counts frames with usable context; the other columns exclude dropped frames. Bold marks the best of the three representations in each column. Rows are ordered by balanced accuracy.}
    \label{tab:traffic}
\end{table}

\section{Prompt Catalog}
\label{app:prompts}

Curly-brace placeholders are filled per request.

\subsection{Traffic Evaluation}
\label{app:prompts-traffic}

\begin{promptbox}{Embedding verdict: traffic prototype prompts (Cosmos-Embed1 text encoder)}
\detokenize{a warehouse aisle blocked by a forklift}\par
\detokenize{a warehouse aisle blocked by a pallet}\par
\detokenize{a warehouse aisle blocked by stacked boxes}\par
\detokenize{a person walking in a warehouse aisle}\par
\detokenize{a mobile robot blocking a warehouse aisle}\par
\detokenize{a warehouse aisle blocked by a pallet jack}
\end{promptbox}

\begin{promptbox}{Embedding verdict: clear prototype prompts (Cosmos-Embed1 text encoder)}
\detokenize{an empty warehouse aisle}\par
\detokenize{a clear, open warehouse aisle}\par
\detokenize{a warehouse aisle with an unobstructed path}
\end{promptbox}

\begin{promptbox}{Embedding verdict: sentence written to memory}
\detokenize{In the aisle straight ahead there is traffic.}\par
\detokenize{In the aisle straight ahead there is no traffic.}
\end{promptbox}

\begin{promptbox}{Detection arm: detection prompt (Cosmos3-Edge)}
\detokenize{Return only one JSON array of objects. Each object must have exactly {"label": string, "bbox_2d": [x_min, y_min, x_max, y_max]} using normalized 0-1000 coordinates. Include distinct whole vehicles or agents physically occupying or narrowing the navigable path directly ahead; use one box per object, never object parts or duplicate boxes. Label each object by its category, such as forklift, vehicle, or person; never use a generic label such as obstacle. Exclude the ego vehicle, railings, floor markings, fixtures, fixed infrastructure, and objects beside the path. Do not include markdown or any text outside the JSON array.}
\end{promptbox}

\begin{promptbox}{Detection arm: correction prompt for the second and third attempts (Cosmos3-Edge)}
\detokenize{Return only one JSON array. Each object must have exactly {"label": string, "bbox_2d": [x_min, y_min, x_max, y_max]} with normalized 0-1000 coordinates. Include only distinct whole vehicles or agents that block or narrow the path directly ahead; exclude object parts, duplicate boxes, railings, floor markings, fixtures, fixed infrastructure, the ego vehicle, and beside-path objects. Label each detection by its object category, such as forklift, vehicle, or person, never obstacle. Do not include markdown or explanatory text.}
\end{promptbox}

\begin{promptbox}{Description arm: description prompt (Cosmos3-Edge)}
\detokenize{This is a single still image for robot navigation. In one concise sentence of at most 40 words, describe the relevant objects, obstacle layout, free space, and navigable paths. Do not infer motion, speed, or events over time.}
\end{promptbox}

\begin{promptbox}{Agent question, one per frame (Nemotron-3-Nano-4B)}
\detokenize{It is {HH:MM:SS} and I am a robot at position {[x, y, z]} in the warehouse. What is the traffic information in this area? Specifically, is there any other vehicle or agent in the zone directly ahead of me right now? Answer with only one word yes or no.}
\end{promptbox}

\subsection{Live Semantic Map}
\label{app:prompts-map}

\begin{promptbox}{Map query, one per robot per poll (Nemotron-3-Nano-4B)}
\detokenize{The time now is {HH:MM:SS}. You are a robot at position {[x, y, z]} in a warehouse. We are looking to know the traffic situation in the position where you are in right now. Is there any other vehicle or agent in the zone directly ahead of you right now? Answer with only one word yes or no.}
\end{promptbox}

\begin{promptbox}{Operator description request on robot selection (Cosmos3-Edge language decoder)}
\detokenize{Provide two sentences on the traffic status of the aisle in front of you. Mention other vehicles or agents in the scene.}
\end{promptbox}

\subsection{Agent Interfaces}
\label{app:prompts-agent}

\begin{promptbox}{Search tools}
\detokenize{retrieve_from_text(x: string)}\par
\detokenize{retrieve_from_position(x: (x, y, z))}\par
\detokenize{retrieve_from_time(x: "HH:MM:SS")}
\end{promptbox}

\begin{promptbox}{Describe tool}
\detokenize{describe(memory_id, question) -> text}
\end{promptbox}

\begin{promptbox}{Final structured answer}
\detokenize{fields: text, binary (yes or no), position, time, duration}\par
\detokenize{position is null or three numbers}
\end{promptbox}

\section{Training on Cached Representations}
\label{app:finetune}

\begin{figure}[H]
    \centering
    \resizebox{\linewidth}{!}{%
        \begin{tikzpicture}[
            font=\sffamily\footnotesize,
            component/.style={draw=black!25, fill=black!4, rounded corners=3pt, text width=2.5cm, minimum height=1.0cm, align=center, inner sep=3pt},
            model/.style={component, draw=nvidiagreen!75!black, fill=nvidiagreen!10},
            memory/.style={component, draw=blue!35!black, fill=blue!5},
            flow/.style={-{Stealth[length=2.1mm]}, draw=black!65, line width=0.7pt},
            note/.style={font=\sffamily\scriptsize, text=black!65, fill=white, inner sep=1.5pt},
            tok/.style={draw, rounded corners=2pt, minimum width=0.72cm, minimum height=0.5cm, inner sep=1pt, font=\sffamily\scriptsize},
            tvis/.style={tok, draw=cyan!50!black, fill=cyan!8},
            tlang/.style={tok, draw=blue!50!black, fill=blue!6},
            tspec/.style={tok, draw=black!35, fill=black!6, font=\sffamily\tiny},
            tdm/.style={tok, draw=orange!70!black, fill=orange!12},
            tact/.style={tok, draw=green!40!black, fill=green!8},
            noisy/.style={dashed},
            noise/.style={circle, draw=black!65, fill=white, inner sep=0pt, minimum size=4.4mm, font=\sffamily\scriptsize}
        ]
            \coordinate (tokrow) at (0,5.68);

            \node[memory, text width=2.6cm] (H) at (1.62,7.4) {\textbf{Cached vision tokens $\mathbf{H}$}\\SeaweedFS};
            \node[component, text width=3.6cm] (tokz) at (5.24,7.4) {\textbf{Language tokenizer}\\Prompt, caption, label (Milvus)};
            \node[memory, text width=2.6cm] (Z) at (10.22,7.4) {\textbf{Cached Wan2.2 latents $\mathbf{z}_0$}\\SeaweedFS};
            \node[component, text width=2.6cm] (act) at (14.42,7.4) {\textbf{Action encoder}\\Aligned action targets};
            \node[note, fill=none, text=black!50] at (1.62,8.35) {\itshape Replaces Vision Encoder (ViT)};
            \node[note, fill=none, text=black!50] at (10.22,8.35) {\itshape Replaces Vision Encoder (VAE)};

            \node[tvis] at (0.80,5.4) {$h_1$};
            \node[tvis] at (1.62,5.4) {$h_2$};
            \node[tvis] at (2.44,5.4) {$\cdots$};
            \node[tlang] at (3.60,5.4) {$l_1$};
            \node[tlang] at (4.42,5.4) {$l_2$};
            \node[tlang] at (5.24,5.4) {$\cdots$};
            \node[tspec] at (6.06,5.4) {EOS};
            \node[tspec] at (6.88,5.4) {BOG};

            \node[tdm, noisy] at (9.40,5.4) {$\tilde{z}_1$};
            \node[tdm, noisy] at (10.22,5.4) {$\tilde{z}_2$};
            \node[tdm, noisy] at (11.04,5.4) {$\cdots$};
            \node[tact, noisy] at (13.60,5.4) {$\tilde{a}_1$};
            \node[tact, noisy] at (14.42,5.4) {$\tilde{a}_2$};
            \node[tact, noisy] at (15.24,5.4) {$\cdots$};

            \draw[flow] (H.south) -- (H.south |- tokrow);
            \draw[flow] (tokz.south) -- (tokz.south |- tokrow);

            \node[noise] (nz) at (10.22,6.25) {$+$};
            \draw[flow] (Z.south) -- (nz.north);
            \draw[flow] (nz.south) -- (nz.south |- tokrow);
            \node[note, anchor=west] at ([xshift=2pt]nz.east) {$\mathbf{z}_t = (1-t)\,\mathbf{z}_0 + t\,\boldsymbol{\epsilon}$};

            \node[noise] (na) at (14.42,6.25) {$+$};
            \draw[flow] (act.south) -- (na.north);
            \draw[flow] (na.south) -- (na.south |- tokrow);

            \draw[draw=blue!50!black, line width=0.6pt] (0.35,5.0) -- (0.35,4.9) -- (7.33,4.9) -- (7.33,5.0);
            \node[font=\sffamily\bfseries\footnotesize, text=blue!50!black] (arl) at (3.84,4.6) {AR subsequence (Reasoner)};
            \draw[draw=orange!70!black, line width=0.6pt] (8.95,5.0) -- (8.95,4.9) -- (15.69,4.9) -- (15.69,5.0);
            \node[font=\sffamily\bfseries\footnotesize, text=orange!70!black] (dml) at (12.32,4.6) {Diffusion subsequence (Generator)};

            \node[model, text width=15.0cm, minimum height=1.2cm] (tf) at (8.02,3.3)
                {\textbf{Cosmos3 Mixture-of-Transformers}\enspace$\times L$ \enspace(same architecture)\\
                 Shared multimodal attention: causal over AR tokens; DM queries attend to $[\mathbf{K}_{\mathrm{AR}}; \mathbf{K}_{\mathrm{DM}}]$};
            \draw[flow] (arl.south) -- (arl.south |- tf.north);
            \draw[flow] (dml.south) -- (dml.south |- tf.north);

            \node[tlang] at (3.60,1.9) {$l_2$};
            \node[tlang] at (4.42,1.9) {$l_3$};
            \node[tlang] at (5.24,1.9) {$\cdots$};
            \node[tdm] at (9.40,1.9) {$\hat{z}_1$};
            \node[tdm] at (10.22,1.9) {$\hat{z}_2$};
            \node[tdm] at (11.04,1.9) {$\cdots$};
            \node[tact] at (13.60,1.9) {$\hat{a}_1$};
            \node[tact] at (14.42,1.9) {$\hat{a}_2$};
            \node[tact] at (15.24,1.9) {$\cdots$};

            \draw[flow] ([xshift=-3.6cm]tf.south) -- ++(0,-0.55);
            \draw[flow] ([xshift=2.2cm]tf.south) -- ++(0,-0.55);
            \draw[flow] ([xshift=6.4cm]tf.south) -- ++(0,-0.55);

            \node[note, anchor=west, fill=none] at (5.70,1.9) {$\rightarrow\ \mathcal{L}_{\mathrm{AR}}$};
            \node[note, anchor=west, fill=none] at (11.50,1.9) {$\rightarrow\ \mathcal{L}_{\mathrm{DM}}$};

            \node[tvis, minimum width=0.5cm] at (0.6,0.8) {};
            \node[note, anchor=west, fill=none] at (0.9,0.8) {Cached vision (AR)};
            \node[tlang, minimum width=0.5cm] at (3.5,0.8) {};
            \node[note, anchor=west, fill=none] at (3.8,0.8) {Language};
            \node[tspec, minimum width=0.5cm] at (5.4,0.8) {};
            \node[note, anchor=west, fill=none] at (5.7,0.8) {Special};
            \node[tdm, minimum width=0.5cm] at (7.1,0.8) {};
            \node[note, anchor=west, fill=none] at (7.4,0.8) {Cached latent (DM)};
            \node[tact, minimum width=0.5cm] at (10.1,0.8) {};
            \node[note, anchor=west, fill=none] at (10.4,0.8) {Action};
            \node[tok, noisy, draw=black!40, minimum width=0.5cm] at (11.7,0.8) {};
            \node[note, anchor=west, fill=none] at (12.0,0.8) {Noisy};
        \end{tikzpicture}%
    }
    \caption{Fleet memory supplies Cosmos3 fine-tuning without rerunning the encoders. Cached vision tokens $\mathbf{H}$ feed the vision-language reasoner, and cached Wan2.2 latents $\mathbf{z}_0$, noised to level $t$, feed the generator. Semantic filtering selects high-priority memories, and validated text labels and aligned action targets supply supervision. With everything upstream of the cache boundary frozen, the language, diffusion or action components can be adapted (Appendix~\ref{app:finetune}).}
    \label{fig:cosmos3-injection}
\end{figure}
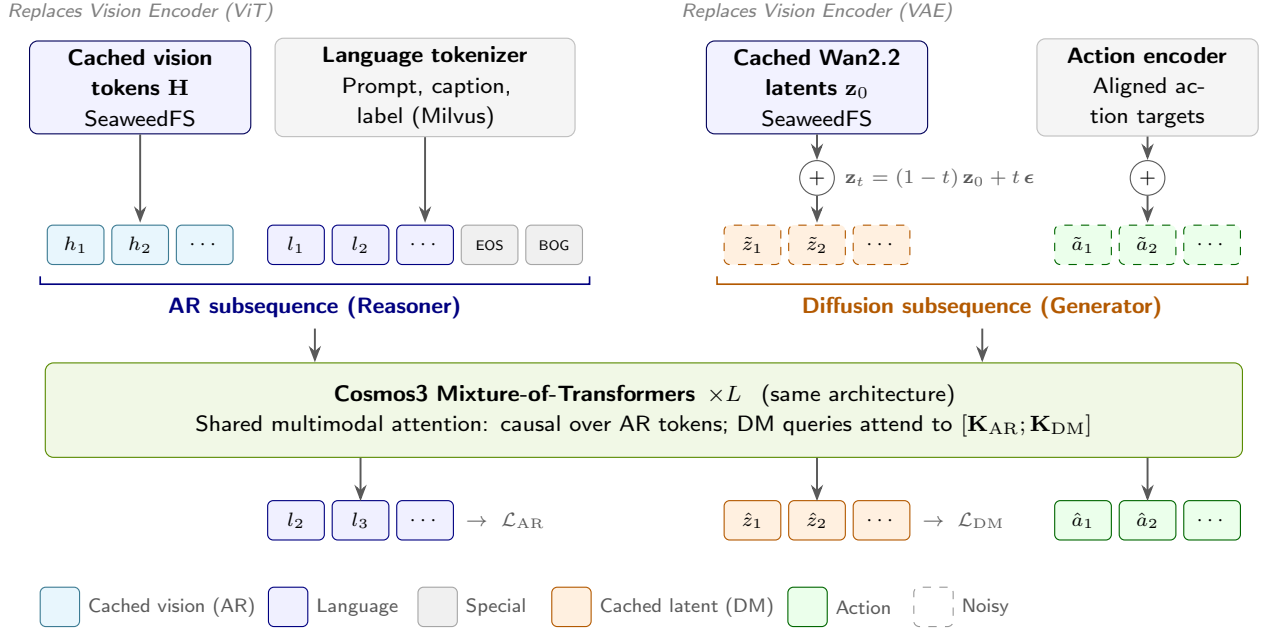

Fleet operation already performs the preprocessing that Cosmos3 training needs. Each memory stores the Cosmos3-Edge vision tokens $\mathbf{H}=f_\phi(\mathbf{x})$ and the Wan2.2 video latents $\mathbf{z}_0=\mathrm{VAE}(\mathbf{x})$~\citep{nvidia2026cosmos3}, so training reads both from the object store without moving or re-encoding video (\FIGREF{fig:cosmos3-injection}). Cosmos-Embed1 queries select relevant events and diverse scenes for the training set. Over $E$ epochs on $M$ memories, the cache saves $EM$ forward passes of each encoder, and training compute goes only to the components being adapted: the language decoder, diffusion components or action heads.

\textbf{Conditions.}
\begin{itemize}
    \item \emph{Frozen cache boundary.} The learnable parameters of everything upstream of a cached tensor remain frozen through finetuning.
    \item \emph{Pinned encoders.} Each artifact records its encoder version, preprocessing and tensor layout, and the consumer must match the Cosmos3 variant and vision interface that produced it.
    \item \emph{Shared latent space.} The world model trains in the Wan2.2 VAE latent space. Another tokenizer needs re-encoding or a learned latent adapter.
    \item \emph{Targets.} Recorded latents are self-supervised targets for the world model. Language training uses operator-validated answers attached to selected memories, and action adaptation needs synchronized, embodiment-compatible action targets.
\end{itemize}

\textbf{Objectives.} The language decoder (parameters $\omega$, or a LoRA adapter) learns to write a validated answer $\mathbf{y}$ to a prompt $\mathbf{q}$ with $\mathbf{H}$ in the video slot. The world model $v_\psi$ is trained by flow matching on a clean latent $\mathbf{z}_0$, with noise $\boldsymbol{\epsilon}\sim\mathcal{N}(\mathbf{0},\mathbf{I})$, time $t\sim\operatorname{Uniform}[0,1]$ and conditioning $\mathbf{c}$ (template text, pose, time and preceding latents):
\begin{align}
    \mathcal{L}_{\mathrm{AR}}(\omega) &= -\,\mathbb{E}_{(\mathbf{H}, \mathbf{q}, \mathbf{y})} \sum_{j} \log p_\omega\!\left(y_j \mid y_{<j},\, \mathbf{q},\, \mathbf{H}\right), \\
    \mathcal{L}_{\mathrm{DM}}(\psi) &= \mathbb{E}_{(\mathbf{z}_0, \mathbf{c}),\, t,\, \boldsymbol{\epsilon}} \left\| v_\psi(\mathbf{z}_t, t, \mathbf{c}) - (\boldsymbol{\epsilon} - \mathbf{z}_0) \right\|_2^2,
    \qquad \mathbf{z}_t = (1 - t)\,\mathbf{z}_0 + t\,\boldsymbol{\epsilon}.
\end{align}

\end{document}